\documentclass[10pt,twocolumn,letterpaper]{article}

\usepackage{iccv}
\usepackage{times}
\usepackage{epsfig}
\usepackage{graphicx}
\usepackage{amsmath}
\usepackage{amssymb,color,multirow}
\def\relu{ReLU\xspace}
\def\loloss{\ensuremath{L_1}\xspace}
\def\ltloss{\ensuremath{L_2}\xspace}
\usepackage{pgfplots}
\pgfplotsset{
  compat=newest,
  xlabel near ticks,
  ylabel near ticks
}

\usepackage[pagebackref=true,breaklinks=true,letterpaper=true,colorlinks,bookmarks=false]{hyperref}

\iccvfinalcopy 
\newif\ifwithlatentalgo
\newif\iflong
\newif\iflettrine
\newif\ifiksvm
\newif\ifaptable
\longtrue
\newif\ifmultitables
\multitablestrue
\newif\ifappendix
\appendixtrue
\newif\iflongtables
\newif\ifltpap
\newif\ifnooverview

\makeatletter
\DeclareRobustCommand\onedot{\futurelet\@let@token\@onedot}
\def\@onedot{\ifx\@let@token.\else.\null\fi\xspace}
\usepackage{xspace}
\usepackage{upquote,textcomp}

\def\bf#1{\textbf{#1}}

\def\ie{\emph{i.e}\onedot}

\def\wrt{\emph{w.r.t}\onedot}

\def\cf{\emph{c.f}\onedot}

\def\etc{\emph{etc}\onedot}

\def\etal{\emph{et al}\xspace}

\def\para#1{{\bf #1.}}

\def\mbold#1{\textbf{#1}}
\def\mitalic#1{\textit{#1}}

\newif\ifwithcomments 
\withcommentstrue
\ifwithcomments

\long\def\comment#1{\para{\sffamily\{***\color{RoyalBlue}#1}****\} }%

\long\def\discuss#1{{\color{ForestGreen}\mbold{\underline{DISCUSS:}}#1} }
\long\def\diagrams#1{\\{\color{Red}\mbold{\underline{Diagram:}}#1} }

\long\def\idea#1{{\\\color{Magenta} \mbold{Idea:}#1}}
\long\def\revcomment#1{Reviewer BMVC:#1 }
\else

\long\def\comment#1{}
\long\def\discuss#1{ }
\long\def\revcomment#1{ }
\long\def\diagrams#1{ }
\long\def\idea#1{}
\fi

\def\l2norm{\mitalic{L2Norm}}

\def\figref#1{Figure~\ref{figure:#1}}
\newcommand{\figrefs}[2]{Figures \ref{figure:#1} and \ref{figure:#2}}
\def\tabref#1{Table~\ref{table:#1}}
\def\secref#1{Sec.~\ref{sec:#1}}

\def\eqref#1{Eq.~(\ref{eq:#1})}

\newcounter{rno}

\definecolor{bg}{rgb}{0.95,0.95,0.95} 

\def\iccvPaperID{2886} 

\ificcvfinal\fi
\begin{document}

\title{Recovering Homography from Camera Captured Documents using Convolutional Neural Networks}
\maketitle
\footnotetext{This work was done when the author was at NUCES-FAST}

\begin{abstract}
	Removing perspective distortion from hand held camera captured document images is one of the primitive tasks in document analysis, but unfortunately no such method exists that can reliably remove the perspective distortion from document images automatically. In this paper, we propose a convolutional neural network based method for recovering homography from hand-held camera captured documents. 

Our proposed method works independent of document's underlying content and is trained end-to-end in a fully automatic way. Specifically, this paper makes following three contributions: firstly, we introduce a large scale synthetic dataset for recovering homography from documents images captured under different geometric and photometric transformations; secondly, we show that a generic convolutional neural network based architecture can be successfully used for regressing the corners positions of documents captured under wild settings; 
 thirdly, we show that \loloss loss can be reliably used for corners regression. Our proposed method gives state-of-the-art performance on the tested datasets, and has potential to become an integral part of document analysis pipeline.
\end{abstract}

\begin{figure}[!ht]
\begin{center}
\begin{tabular}{c@{\hskip 3pt}c@{\hskip 2pt}c@{\hskip 2pt}c}
\raisebox{+3.5\normalbaselineskip}[0pt][0pt]{\rotatebox[origin=c]{90}{Skewed}}
&
\includegraphics[height=3.5cm]{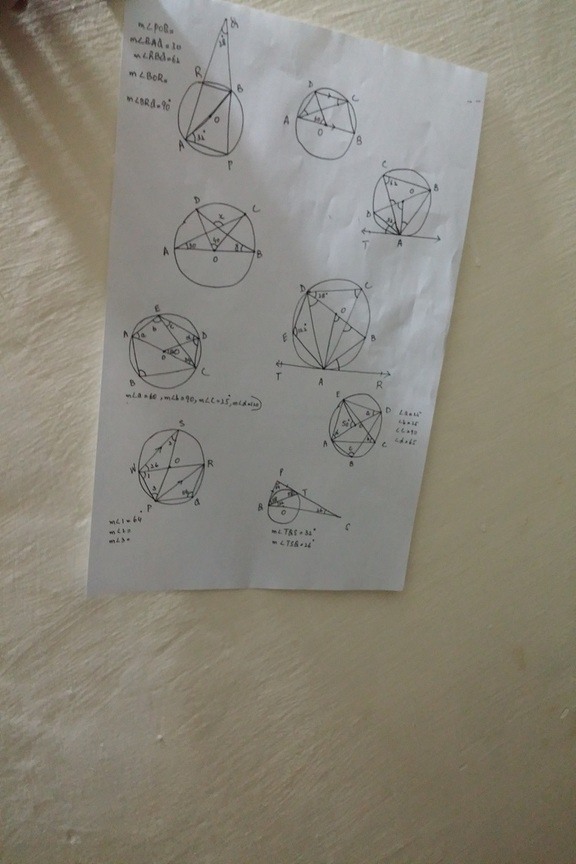}&
\includegraphics[height=3.5cm]{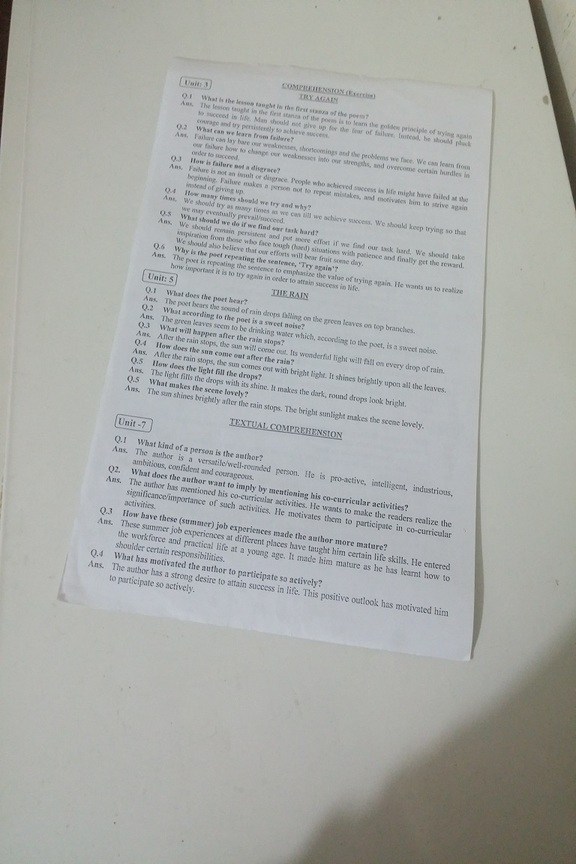}&
\includegraphics[height=3.5cm]{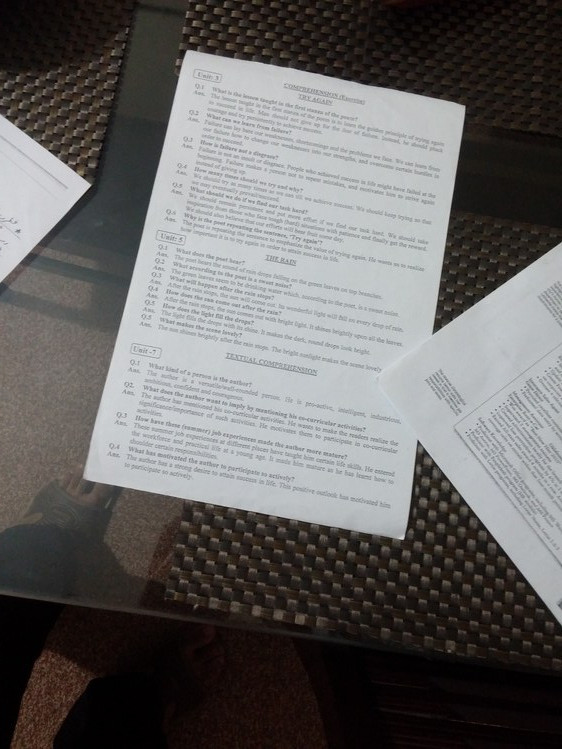}\\
\raisebox{+4\normalbaselineskip}[0pt][0pt]{\rotatebox[origin=c]{90}{De-Skewed}}
&
\includegraphics[height=3.5cm]{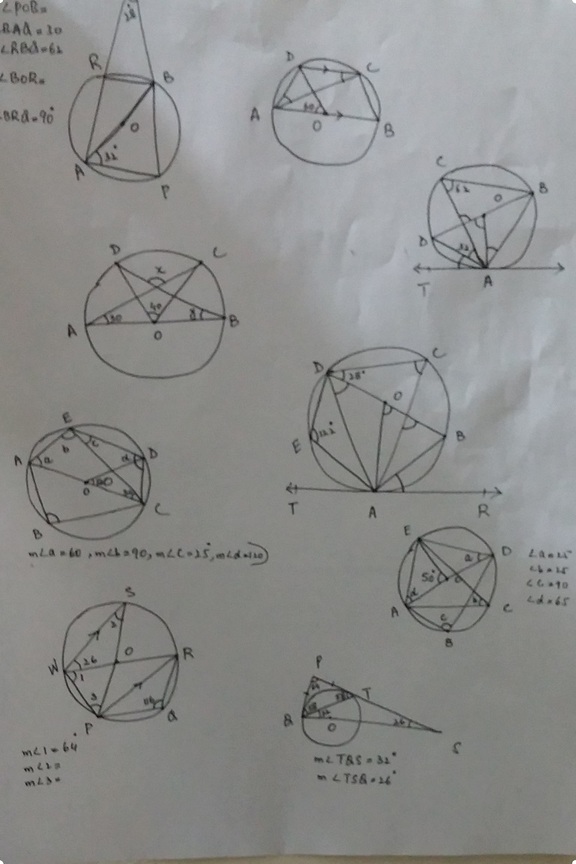}&
\includegraphics[height=3.5cm]{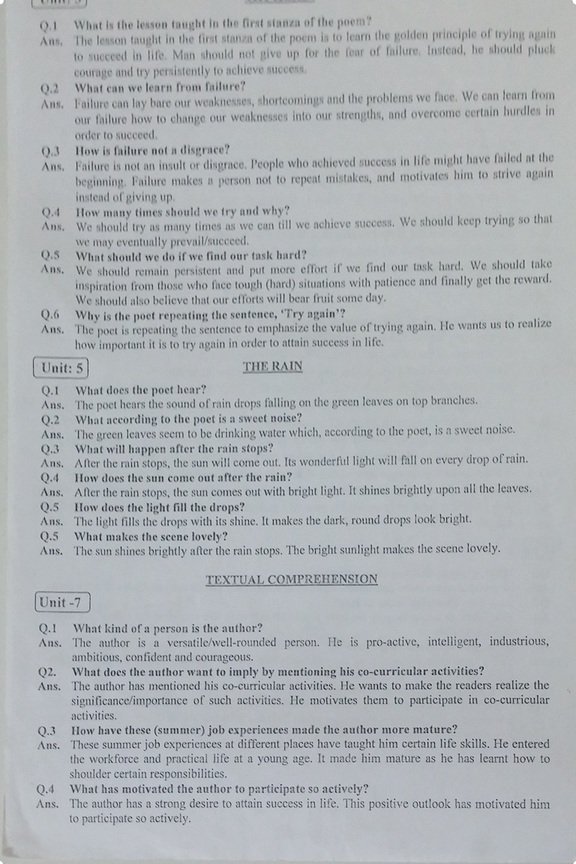}&
\includegraphics[height=3.5cm]{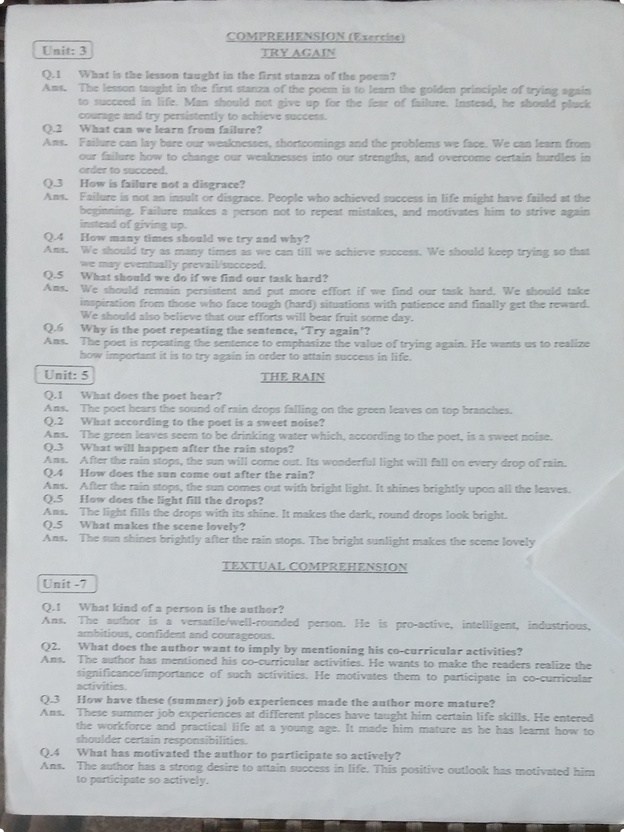}
\end{tabular}

\caption{Results of our method on a set of real world images.} \label{figure:introductory}
\label{figure:examples}
\end{center}
\vspace*{-2em}
\end{figure}

\section{Introduction}
\label{sec:intro}

Hand-held cameras and smart-phones have become an integral part of our today life, and so do the images captured using these devices.  A sizable amount of these images contain textual content and thus requires automatic document analysis pipeline for text detection and understanding. Applying perspective correction to these perspectively warped documents is an important preprocessing step before proceeding with more advanced stages of document analysis pipeline such as binarization, segmentation and optical character recognition.  

Images captured using hand-held devices in real world are significantly different than those captured using dedicated hardwares (such as scanners) in controlled environments, due to the challenging photometric and geometric transformations these images undergo in wild settings. Therefore, traditionally developed methods \cite{ulges2004document,zhang2004restoration,jagannathanperspective} for perspective correction either completely fail or give poor performance, and if not corrected manually (as in majority of commercial applications\footnote{Like \url{www.abbyy.com} and \url{www.camscanner.com}.}) these errors can led to failure of complete document analysis pipeline. 

This degradation in performance can be mainly attributed to the manual and complex pipelines used to estimate the geometric transformation matrix. These pipelines mainly follow the similar setup, where initially low level features such as lines orientation, edges, contours, \etc are used to estimate the page layout and then in later stages, images are deskewed via estimated similarity or homography transformation matrices. However, these methods completely fail in the wild settings due to challenging lighting conditions, cluttered background, motion blur, \etc, -- \cf \secref{software_apps}.

To this end, we propose a generalized and completely automatic method, trained end-to-end, for perspective correction of document images captured in the wild.  

We make following main contributions. Firstly, we introduce and publicly release\footnote{We will release the link to dataset.} a large scale synthetic dataset
for learning homography matrix for perspective correction of textual images in the wild settings. Secondly, we introduce a convolutional neural network based architecture for recovering homography matrix using four points parameterization \cite{baker2006parameterizing} from a single input image. Thirdly, we empirically show that \loloss loss function performs better for homography estimation.

Since our method does not use hand crafted features and is trained over a large enough dataset to capture the different range of photometric and geometric transformations, it works independent of text layout assumptions such as availability of page margins, parallel textlines, \etc in the captured image. Also in comparison to earlier methods, our method is more robust, works under different lighting conditions and presence of different noises, as illustrated by our results. It even works on the occluded documents where significant portion of document is either missing or occluded.  Overall, our method gives state-of-the-art performance on tested datasets. \figref{introductory} show some sample results, please refer to \secref{exp_and_results} for detailed results. 

In addition to being robust, our method is quite simple and relatively fast to train and test. Precisely, it requires around 5 hours for training and requires 0.04s sec for complete forward pass and perspective correction on a Tesla K40 GPU machine.  

The rest of the paper is organized as follows. \secref{rel_work} reviews the related work, while \secref{synthetic_dataset} provides details on our synthetic dataset. \secref{method} explains our CNN model and architecture. \secref{exp_and_results} discusses in detail different experimental choices, parameter settings and our results. Finally, \secref{conclusion} concludes the paper with relevant discussion.
\section{Related Work}
\label{sec:rel_work}
Traditionally, to find homography transformation between a pair of reference and transformed images, first a set of corresponding features is built and then based on this set  either direct linear transform \cite{hartley2003multiple} or cost based methods have been used to estimate the projective transformation \cite{dubrofsky2009homography}.  This step is followed by other post-processing steps to remove false matches or outliers. Over the years, researchers have used corner points, lines, and conics for defining correspondences between pair of images.  However, the whole pipeline is dependent on the quality of detected feature sets and their repeatability, where false correspondences or lack of quality correspondences can lead to large errors in computed transformation matrix. 

For the problem at hand, the above mentioned pipeline cannot be directly used due to absence of reference image. Although a canonical image with white background can be used as reference image however absence of text in the canonical image can lead to misfiring of corner detectors and indirectly to failure of complete methodology. 

In contrast, in document analysis different manual pipelines have been used to restore a perspectively distorted image. We can broadly classify these approaches into two classes. First class of methods \cite{zhang2004restoration,lampert2005oblivious,ulges2004document} make assumptions about the image capturing process to recover the transformation matrix. For instance, Zhang \etal \cite{zhang2004restoration} develop a method for 3D reconstruction of the paper from the shading information in a single image. For this purpose they use special hardware consisting of light sources and sensors.

In contrast, second class of methods make assumptions about document layout \cite{jagannathanperspective, liang2008geometric, shafait2007document,simon2015correcting,bukhari2009dewarping}.  For example, Jagannathan \& Jawahar \cite{jagannathanperspective} extract clues about document layout such as document boundaries, text orientation, page layout information, \etc to either impose constraints for solving the system of linear equations or finding vanishing points and lines for homography estimation. Liang \etal \cite{liang2008geometric} performs projection profile analysis for detecting the orientations of text lines. These text lines orientations are then used for the identification of vanishing points which are then used for the estimation of affine matrix. In 2007, Shafait \etal \cite{shafait2007document} launched a competition in Camera-Based Document Analysis and Recognition Conference (CBDAR) to evaluate different image dewarping algorithms on a standard dataset. In this competition, coordinate transform model produced the best results among the three entries. This method also uses the principle of text lines detection for the rectification of the document. Another method \cite{bukhari2009dewarping} was also applied on the dataset later using ridges based coupled snakes model which obtained even better results. However,  all these methods have same limitations and fail when applied to real world images of captured documents. This is due to the fact that the CBDAR dataset does not contain enough variations to capture the distribution of real world examples. 

Recently, Simon \etal \cite{simon2015correcting} has proposed another method for dewarping of document images. Their developed complex pipeline for perspective rectification involves binarization, blobs and lines detection, and application of morphological operators to  produce a final perspectively rectified image. 

Almost all of these above discussed methods work on images captured in controlled settings with considerable textual cues such as lines, \etc. In addition, these methods use hand crafted features, involve tuning of multiple parameters and are not robust to variations and noises introduced during images captured in wild settings. 
In comparison, our proposed CNN based method can reliably estimate homography from perspectively distorted images without making any assumptions about image content or capturing environments. Recently~\cite{detone2016deep}  have also successfully used convolutional neural networks to estimate homography transformation between pair of natural scene images. Since their method requires a pair of reference and transformed images as input to the network for homography estimation, thus cannot directly work on textual images. In contrast, our method estimates homography from a \emph{single input image} without the need of reference image.

\begin{figure*}[!htb]
\begin{center}
\begin{tabular}{c@{\hskip 2pt}c@{\hskip 2pt}c@{\hskip 2pt}c@{\hskip 2pt}c@{\hskip 2pt}c}
\includegraphics[height=4cm]{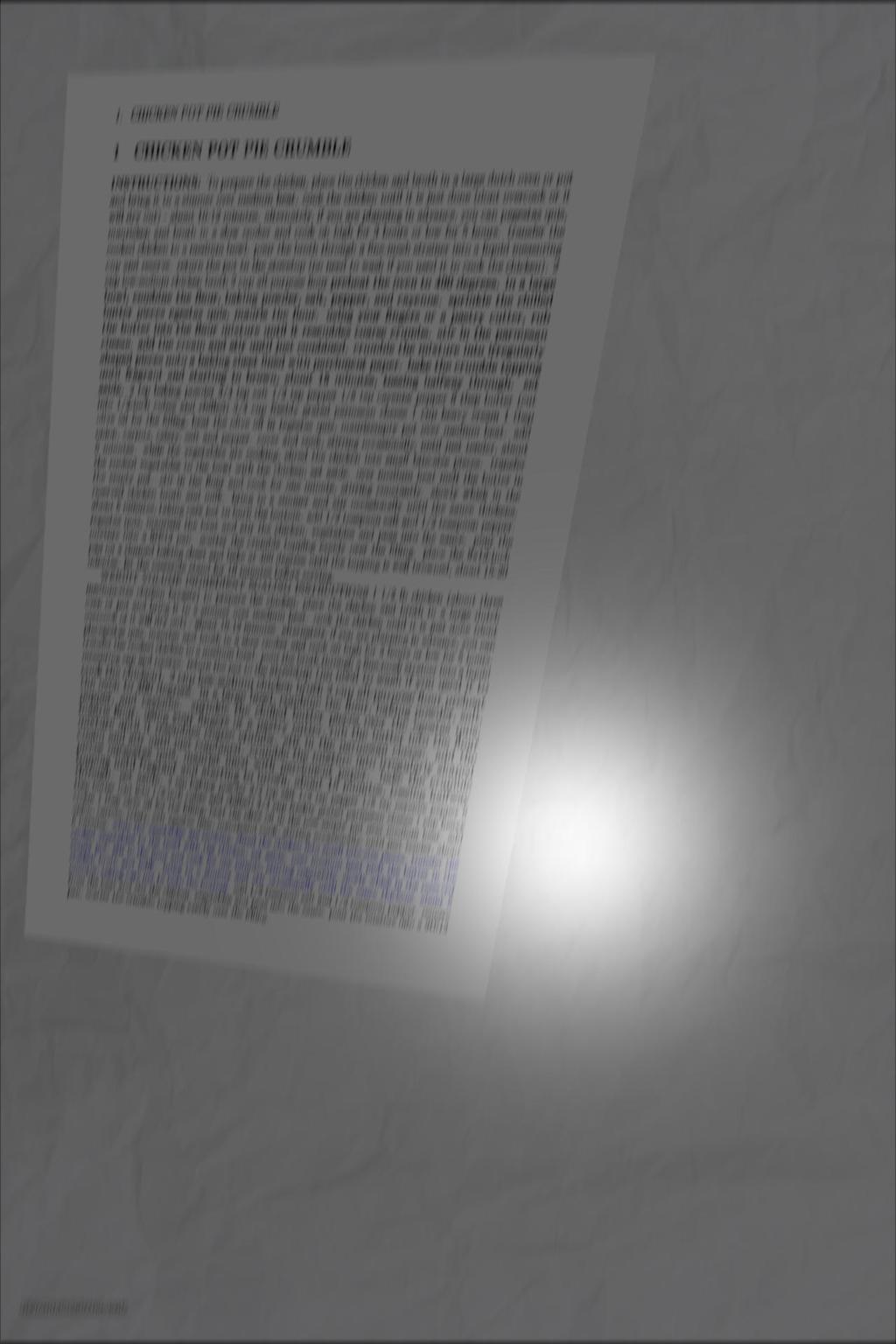}&
\includegraphics[height=4cm]{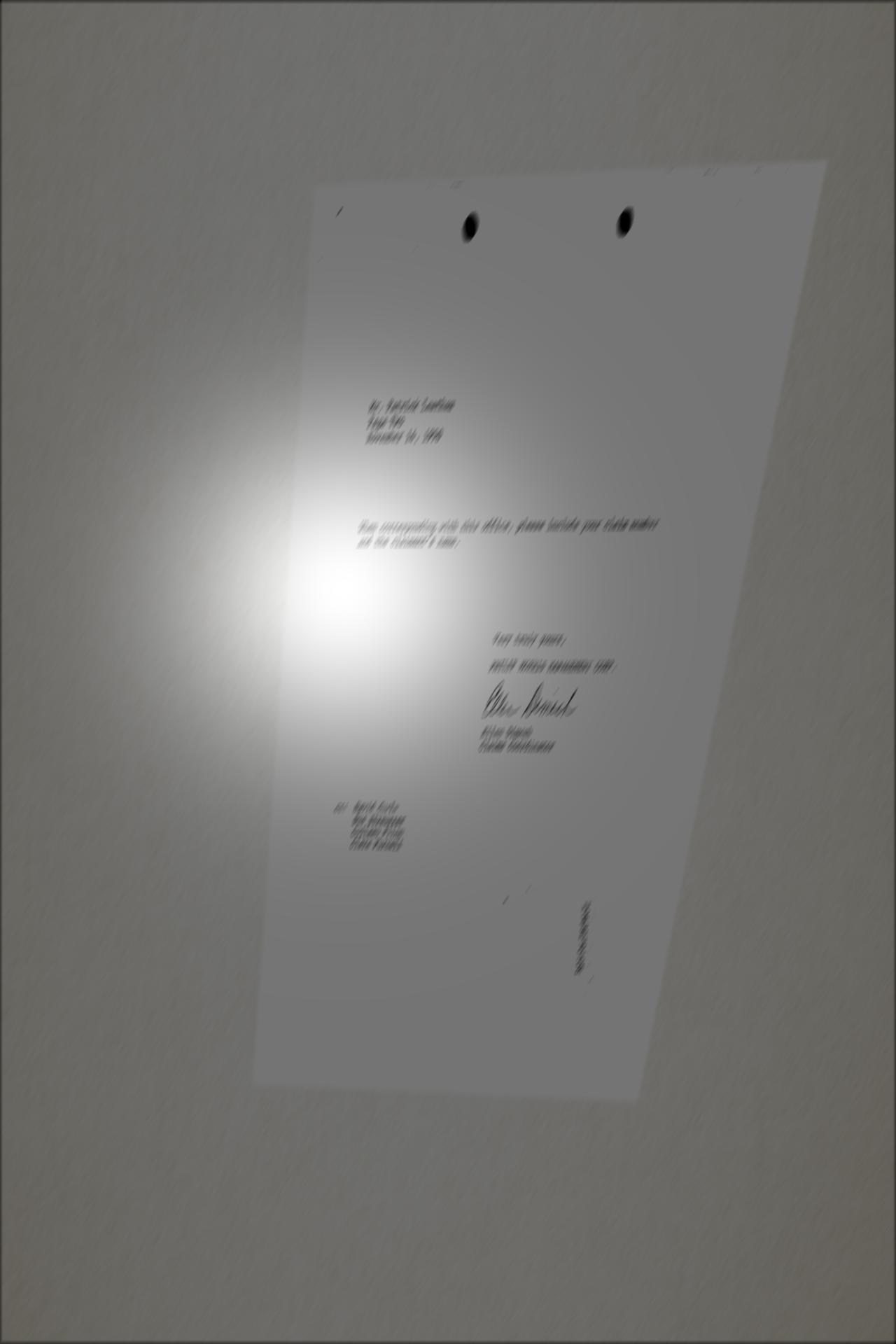}&
\includegraphics[height=4cm]{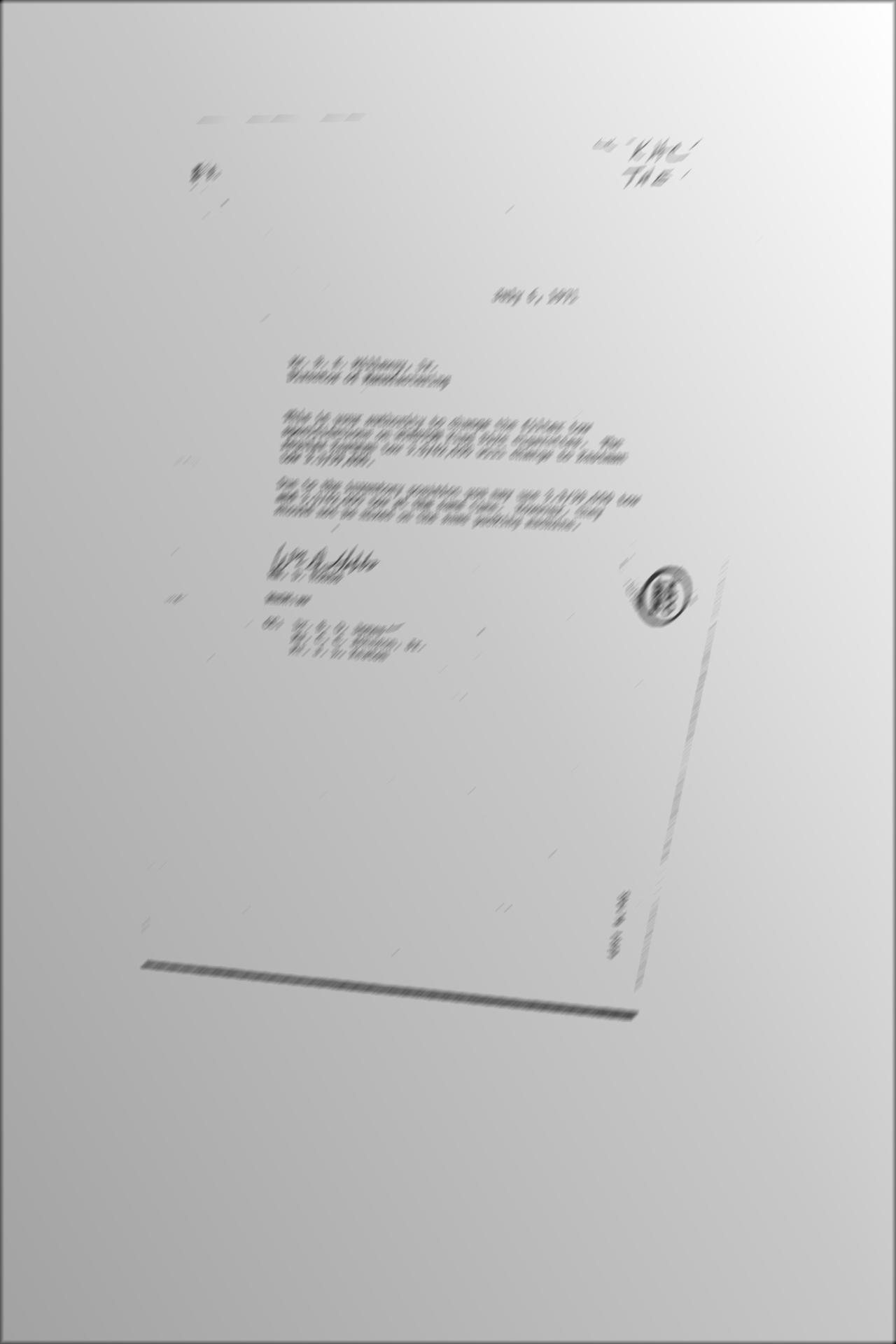}&
\includegraphics[height=4cm]{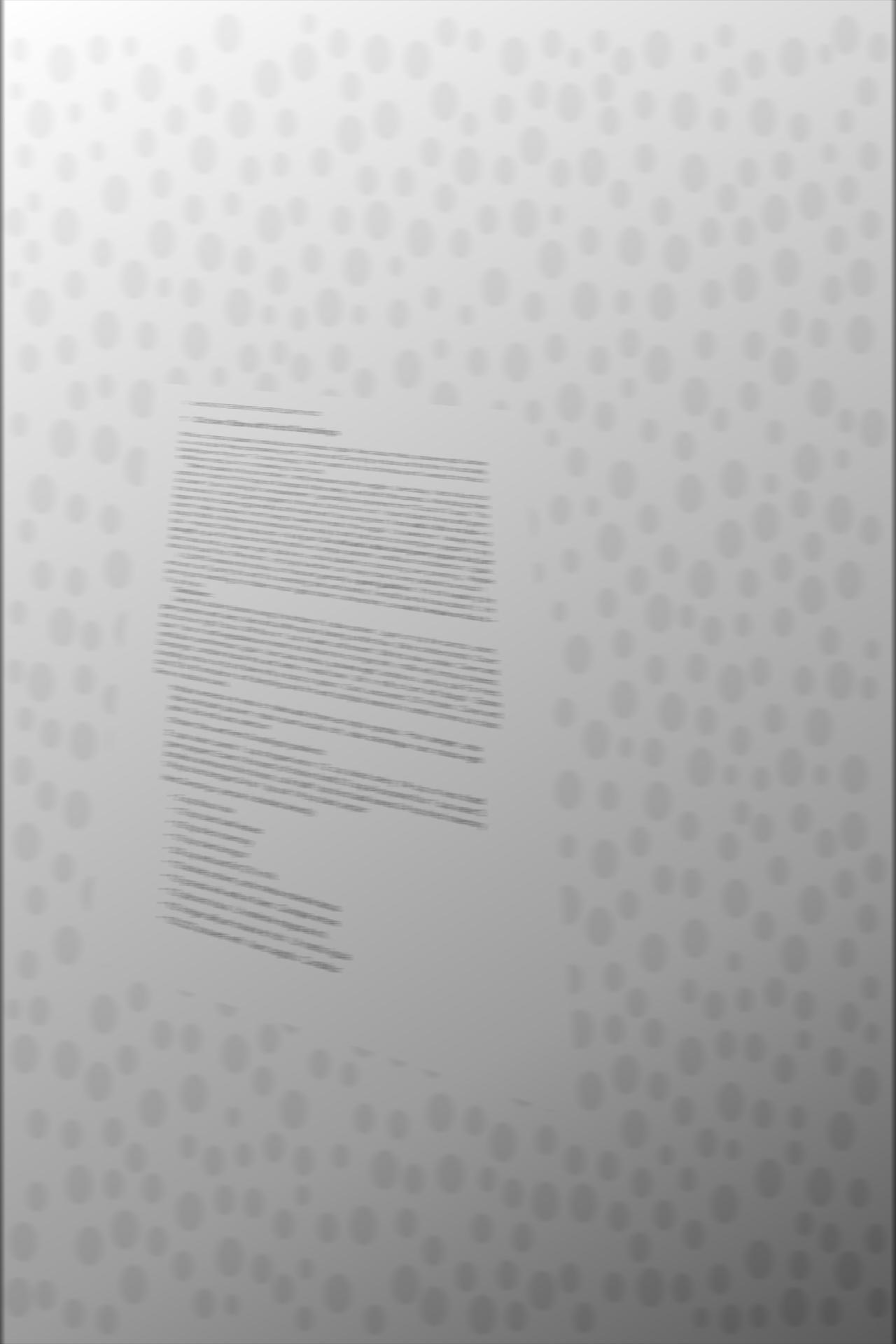}&
\includegraphics[height=4cm]{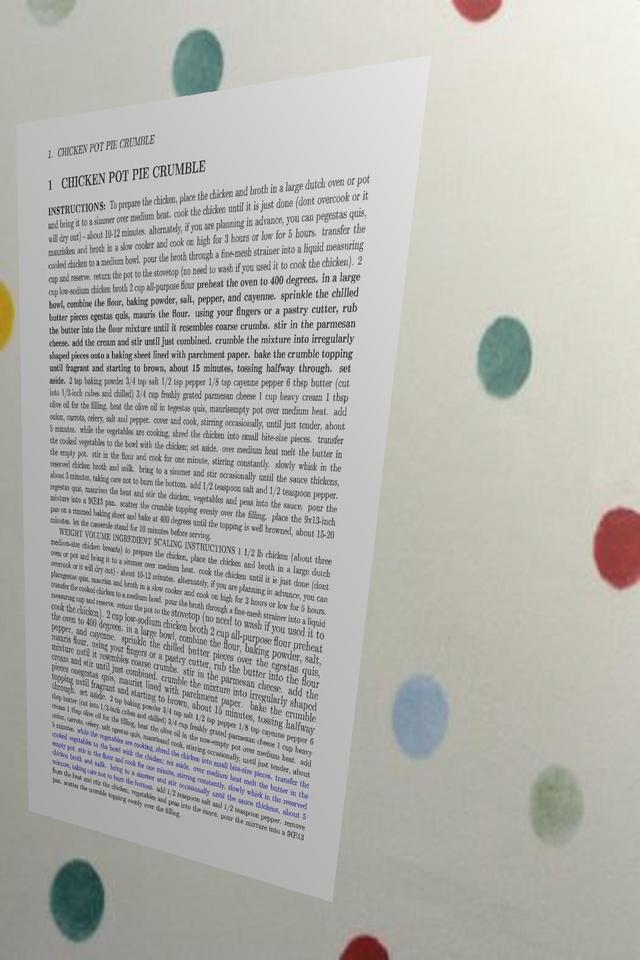}&
\includegraphics[height=4cm]{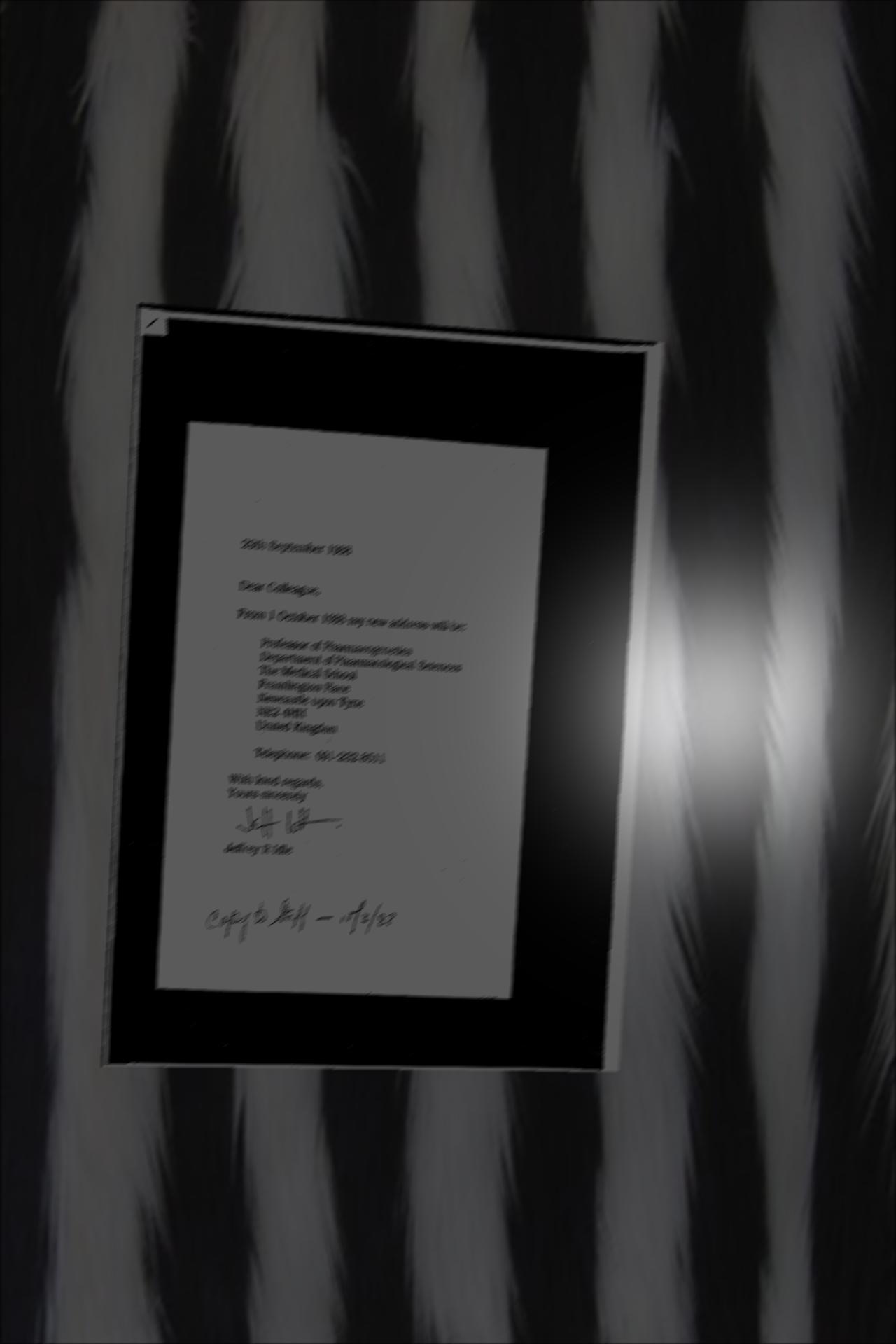}
\end{tabular}
\caption{Sample images from synthetic dataset.}
\label{figure:example_dataset}
\end{center}
\vspace*{-2em}
\end{figure*}
\vspace*{-0.5em}
\section{Synthetic Dataset Generation}
\label{sec:synthetic_dataset} 
Availability of large scale datasets (such as ImageNet~\cite{deng2009imagenet}) has played a significant role in the recent upsurge of deep networks. Unfortunately, as discussed above, no large enough dataset is available for the problem at hand. Currently, the largest available public datasets are CBDAR 2007, CBDAR 2011 \cite{shafait2007document}, and SmartDoc-QA \cite{nayef2015smartdoc}. These datasets either contain very few images or have very limited amount of variability in their images. For instance, CBDAR 2007 and CBDAR 2011 contain around 100 grayscale images of books and periodicals, with limited variations. SmartDoc-QA dataset although contains large number of document images captured under different conditions, these conditions are quite limited compared to wild-settings -- \cf \secref{sdoc}. However, to build a generic image rectification algorithm we need a large dataset of RGB images with large variations in illumination conditions, background cluttering, geometric transformations, \etc, to model the distribution of real world image capturing conditions. 

Recently, researchers have produced and used synthetic datasets to solve data scarcity problem \cite{peng2015learning, gupta2016synthetic, detone2016deep}. Peng \etal~\cite{peng2015learning} build a synthetic dataset to train deep CNNs for learning deep object detectors from 3D models due to limitations of the available dataset. Gupta \etal~\cite{gupta2016synthetic} also train a CNN on a synthetic dataset to solve the problem of text localization in natural images for the same reason. Motivated by the success of these methods, we have developed our own synthetic dataset for the problem at hand.

We use the 3000 document images captured using hand held cameras~\cite{hradivs2015convolutional}\footnote{This dataset was graciously donated by authors.} for building our synthetic dataset. These documents images contain different types of textual content such as text, figures, equations, \etc, and thus serve ideally to capture content variations in the dataset.
\vspace*{-1.5em}
\paragraph{Geometric Transformations:} We first apply different random geometric transformations on these documents to produce perspectively distorted documents. We sample different values of homography matrix $H$ coefficients from a uniform distribution with different ranges.
\begin{equation}
H=
\begin{bmatrix}
  h_{11} & h_{12} & h_{13} \\
  h_{21} & h_{22} & h_{23} \\
  h_{31} & h_{32} & h_{33}
 \end{bmatrix}.\label{eq:H}
\end{equation}

Precisely, $h_{11}$ and $h_{22}$ are randomly sampled from $0.7$ to $1.3$, $h_{12}$ and $h_{21}$ from -$0.3$ to $0.3$, and $h_{31}$ and $h_{32}$ from the range -$0.0015$ to $0.0015$. We further add variable length horizontal and vertical margins to these images to give them camera captured appearance.
\vspace*{-1.3em}
\paragraph{Background Variations:} To introduce the background clutter and variations, as a next step, we add randomly sampled textured backgrounds to these images. For this purpose, we use the Describable Textures Dataset (DTD)~\cite{cimpoi14describing} which contains over 5000 textures from different categories like fibrous, woven, lined, \etc. We have also used Brodatz dataset~\cite{abdelmounaime2013new} which contains 112 textures of different colors and patterns. However, we show in section~\ref{subsec:diffdata} that simple textures alone are not enough to represent the variety of backgrounds that appear in camera captured documents. 

As it turns out, the document images are mainly captured in indoor environments consisting of more complex backgrounds than simple textures. Thus, to model complex indoor backgrounds, we also use MIT Indoor scenes dataset~\cite{quattoni2009recognizing} to sample backgrounds for synthetic images.

\vspace*{-1.3em}
\paragraph{Photometric Transformations:} Images produced using above pipeline appear as real as captured using hand-held cameras but they still lack illumination variations and different noises (such as motion and defocus blur) encountered while capturing images in the wild. To this end, we add motion blur of variable angles and magnitudes to the resultant images to simulate camera shaking and movements effects. We also add Gaussian blur to the images to model dirty lenses and defocus blur. To introduce  different lighting variations, we create different filters based on gamma transformations in spatial domain (we use gamma transformation as a function of displacement from a randomly sampled image position $(x_r, y_r)$ instead of pixel intensity, \ie $I(i,j)=I(i,j)\times ((i-x_r)^2+(j-y_r)^2)^\gamma $) of variable pixel intensities in different directions and shapes. Next we use alpha blending with alpha uniformly sampled from $0.3$ to $0.7$ to merge these filters with the geometrically transformed image. This results in introduction of effects of different lighting variations in the resultant image. Some sample images from our synthetic dataset are shown in figure~\figref{example_dataset}.
\vspace*{-0.5em}
\section{Proposed Method}
\label{sec:method}

\begin{figure*}[!t]
\begin{center}
 \includegraphics[width=0.9\linewidth]{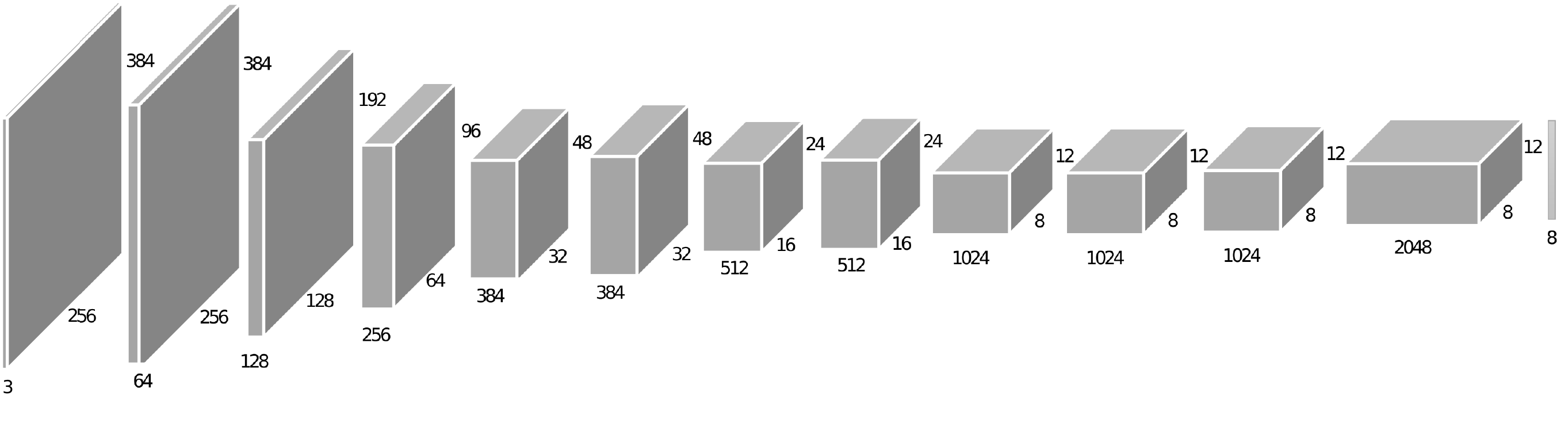}
 \caption{Our architecture consists of 11 convolutional layers and a fully connected layer, initial two layers use filters of size $5\times5$, all of the remaining layers except the last one use filters of size $3\times3$ whereas final layer uses filters of size $1\times1$. The fully connected layer uses 8 neurons to regress the corner positions.}\label{figure:architecture}
 \end{center}
 \vspace*{-2em}
\end{figure*}

In this section, we introduce our convolutional neural network architecture. We experimented with different design choices (such as number of layers, filters, nonlinearities) for our CNN architecture before arriving at final architecture. \figref{architecture} shows our final architecture.

Our final architecture consists of 11 convolutional and maxpooling layers and draws inspirations from the VGG~\cite{simonyan2014very} and FAST YOLO architectures~\cite{yolo2015}. We use \relu nonlinearity after each convolutional layer except the last layer of $1\times1$ convolutions. We use $2\times 2$ max pooling layer after each of the first three, $5_{th}$, and $7_{th}$ layer. The initial two layers use filters of size $5\times5$ whereas all the remaining layers except the last one use filters of size $3\times 3$. The final convolutional layer uses $1\times1$ filters for efficient computation and storage. This is followed by a final fully connected regression layer of $8$ neurons. We use a dropout with $p=0.5$ after the last convolutional layer.

For our loss function, we use \loloss distance to measure the displacement of eight corner coordinates from their canonical positions, \ie
\begin{equation*}
 Loss = \frac{1}{N} \sum_{i=1}^{N} \sum_{j=1}^{4} |y_p^{ij} - y^{ij}|
\end{equation*}

Here $y_p^{ij}$ and $y^{ij}$ represent the $j_{th}$ point predicted and original coordinate values respectively. As mentioned earlier, this formulation is similar to four points homography formulation of~\cite{baker2006parameterizing}.
\vspace*{-.5em}
\section{Experiments and Results}
\label{sec:exp_and_results}

We randomly split our synthetic dataset into three datasets: training, validation and testing. All the configurations have been done using validation set and we report the final performance on the test set.

In literature, researchers \cite{detone2016deep,shafait2007document} have used different metrics to measure the performance of their methods.  For instance,~\cite{detone2016deep} used mean average corner error which is measured using \ltloss distance between the estimated and original corners positions. In comparison,~\cite{shafait2007document} used mean edit distance to compare the methods on the CBDAR 2007 dataset. In our experiments, we use Mean Displacement Error (MDE) by computing the average of \loloss distance between ground truth corner coordinates and the predicted corner coordinates of a document, this measure is then averaged over complete dataset to get the final score for the dataset. In comparison to \ltloss distance, MDE gives better intuition and insights (as we can directly know in pixel units how well the system is performing) into the performance of the algorithm for the given problem.
\vspace*{-0.5em}\subsection{Implementation Details}\label{subsec:implementation-details}
We have implemented our system using TensorFlow~\cite{abadi2016tensorflow}. Although our synthetic dataset is composed of different resolutions images, for our network training and evaluation we use fixed size images of $384 \times 256$ resolution, this helps us to train our network with limited resources and employing multiple max pooling layers.
For initialization of our networks we use He \etal \cite{he2015delving} initialization scheme  for \relu based CNN. We use a batch size of 4 during training. We use Adam's Optimization method~\cite{kingma2014adam} with default parameters to train our networks. We set the initial learning  rate to $5E-4$ and reduce it by half whenever the loss stops decreasing. We repeat this reduction process until the absolute change in loss is very small over a few hundred iterations. We also experimented with RMSProp~\cite{tieleman2012lecture} as a choice of optimization method but in our experiments, Adam consistently gave better performance. 

Training our method for around 10 epochs takes on average five hours. During testing, our method takes on average 0.04 seconds on a GPU (NVIDIA Tesla K40) machine per image which translates to roughly 25 images per second.
\subsection{$3\times3$ Homography vs 4-Points Estimation Method}
\label{subsec:3x3 Homography vs 4-point Method}
Initially, we trained our convolutional neural networks to directly predict the $3 \times 3$ homography matrix $H$ (\cf \eqref{H}) as the output. However, these CNN were not able to produce the desired results and were difficult to train. The reason is that the homography matrix $H$ is extremely sensitive to the $h_{31}$  and $h_{32}$ values. That is, even a change in the order of $10^{-3}$ to the $h_{31}$ and $h_{32}$ values results in an incomprehensible resultant image. Later on, we adopted the 4-points method to recover homography from the input image. CNNs trained using this method were more robust to errors in coordinate values, gave much better results and were relatively easier to train than the direct approach.

\subsection{Evaluation of Different Architectures}
\label{subsec:diffarch}
\begin{figure}[!t]
\begin{center}
\includegraphics[width=\linewidth]{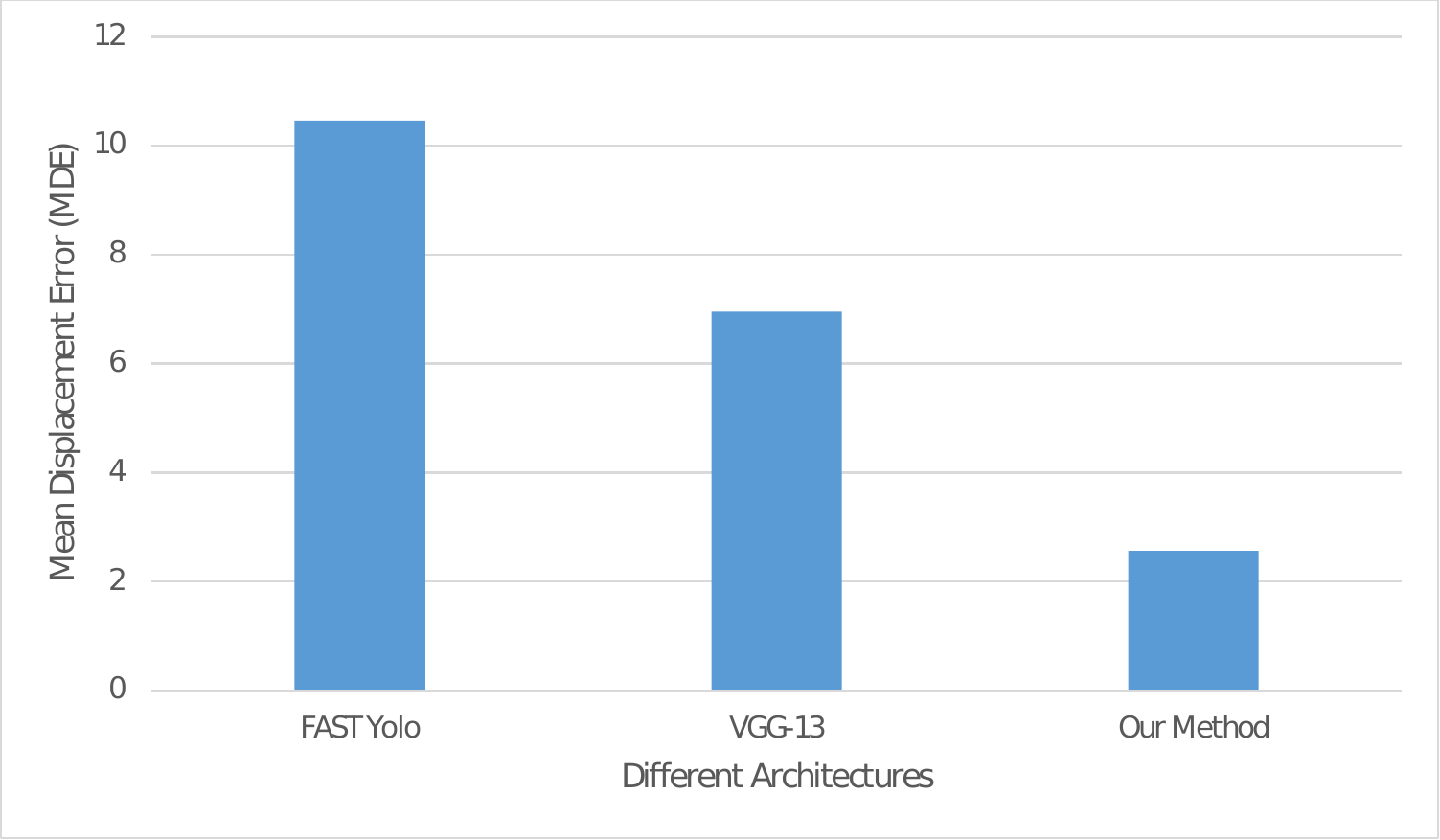}
\end{center}
   \caption{MDE for different architectures on test set.}
\label{figure:histarch}
\vspace*{-1em}
\end{figure}

Initially, We transfer learned a CNN from VGG-13~\cite{simonyan2014very} by replacing the last layer with our prediction layer. This network obtained 6.95 MDE on the test set. We also trained a variant of FAST YOLO~\cite{yolo2015}. This variant reported an MDE of 10.46 pixels on the test dataset.  In our initial analysis, we found out that having large number of filters with large receptive fields in initial layers plays a critical role in the performance of our system. This is because large filters with large receptive fields  are able to capture local co-occurrence statistics much better in the initial layers.  Secondly, we found out that going deeper leads to much better results.  Therefore, based on these findings we designed our own architecture as already discussed in \secref{method}. Our this final architecture includes more number of filters and convolutional layers as compared to YOLO and large receptive fields as compared to VGG. Our this architecture was able to obtain state of the art MDE of 2.45 pixels on our test set -- \cf~\figref{histarch}.

\figrefs{examples}{software_comp} show the results of our proposed architecture on unseen real world images.

\begin{figure*}[!ht]
\begin{center}
\begin{tabular}{c@{\hskip 2pt}c@{\hskip 2pt}c@{\hskip 2pt}c@{\hskip 2pt}c@{\hskip 2pt}c}
\includegraphics[height=4cm]{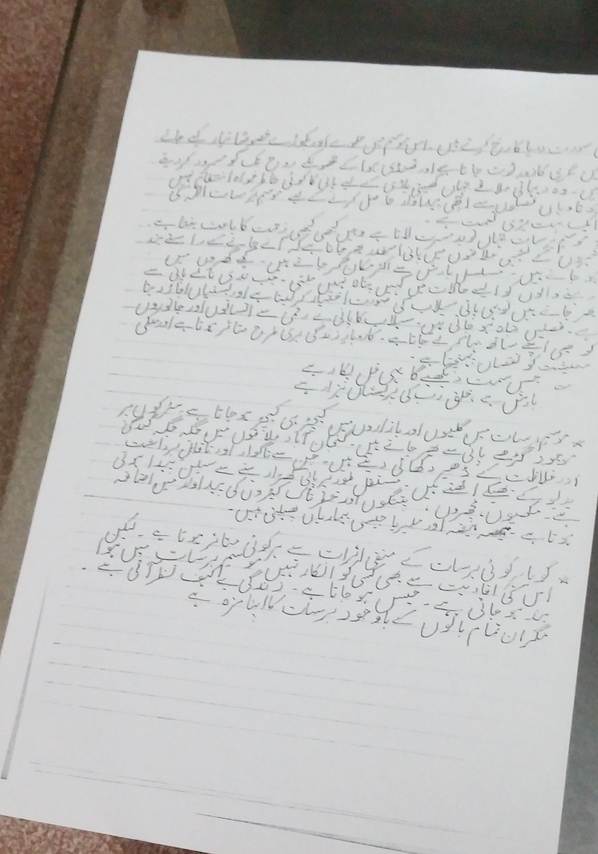}&
\includegraphics[height=4cm]{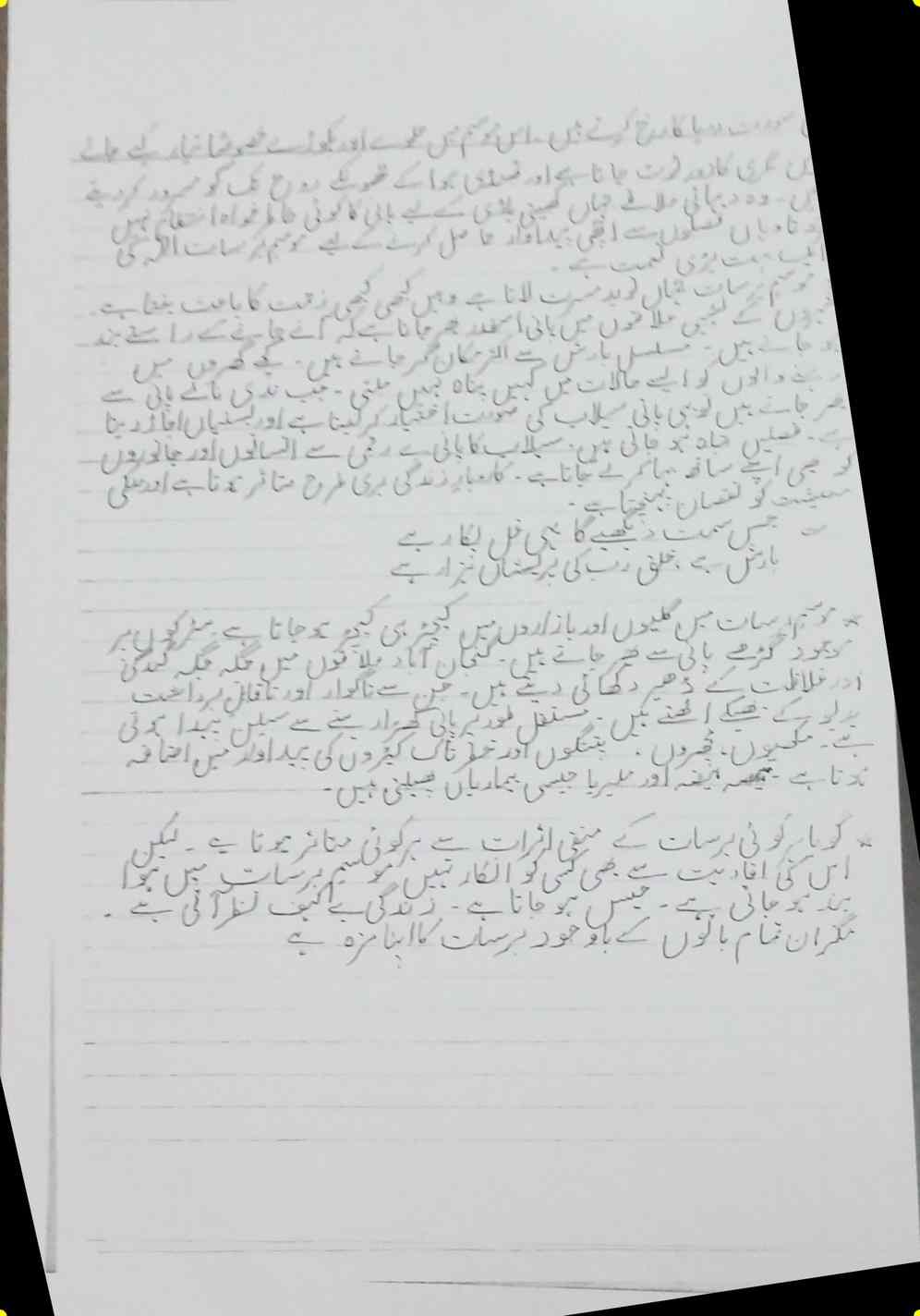}&
\includegraphics[height=4cm]{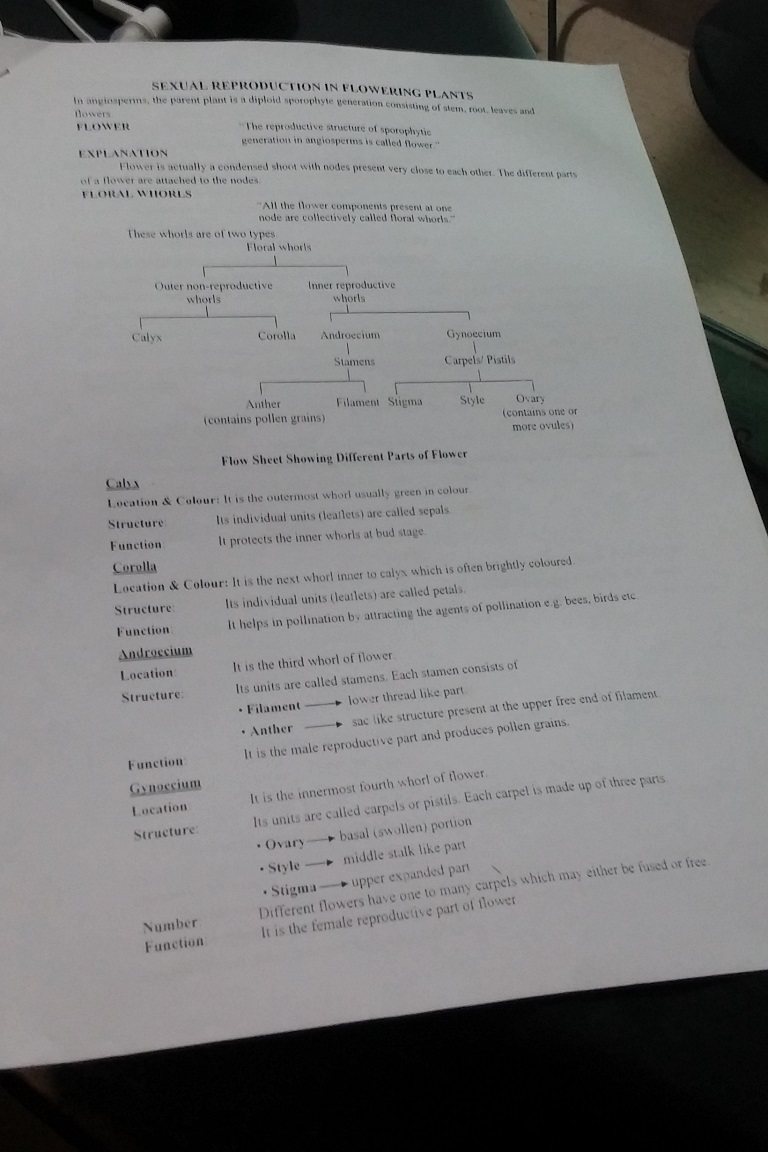} &
\includegraphics[height=4cm]{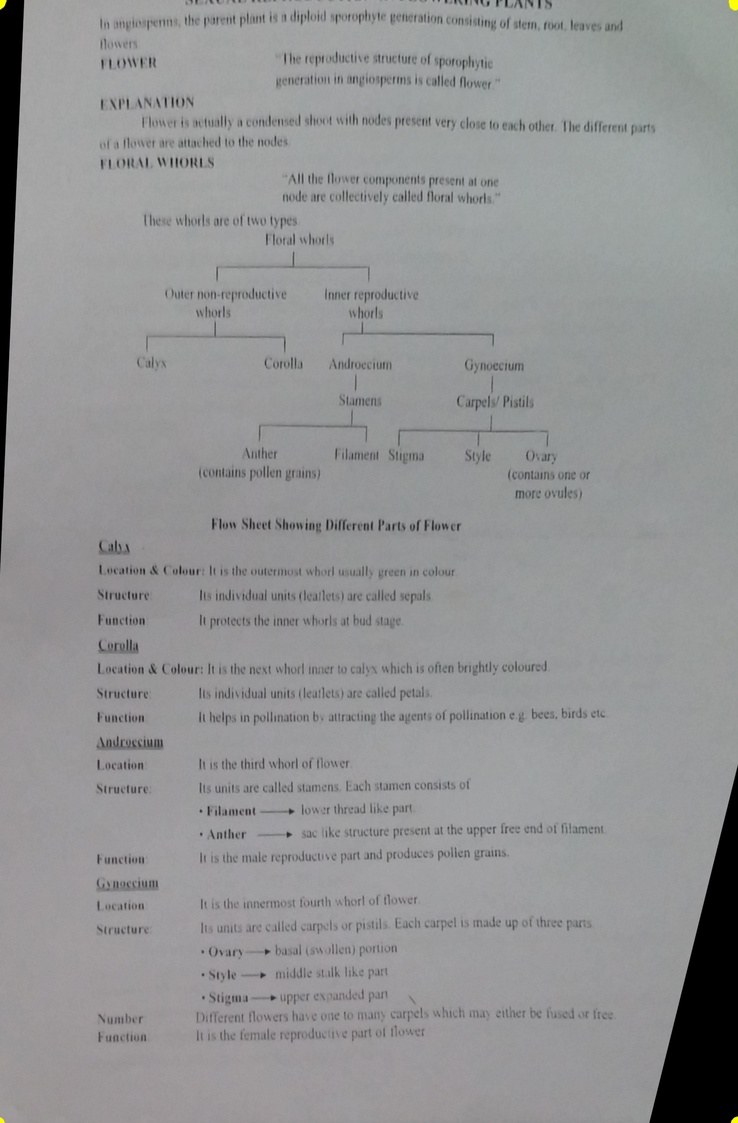} &
\includegraphics[height=4cm]{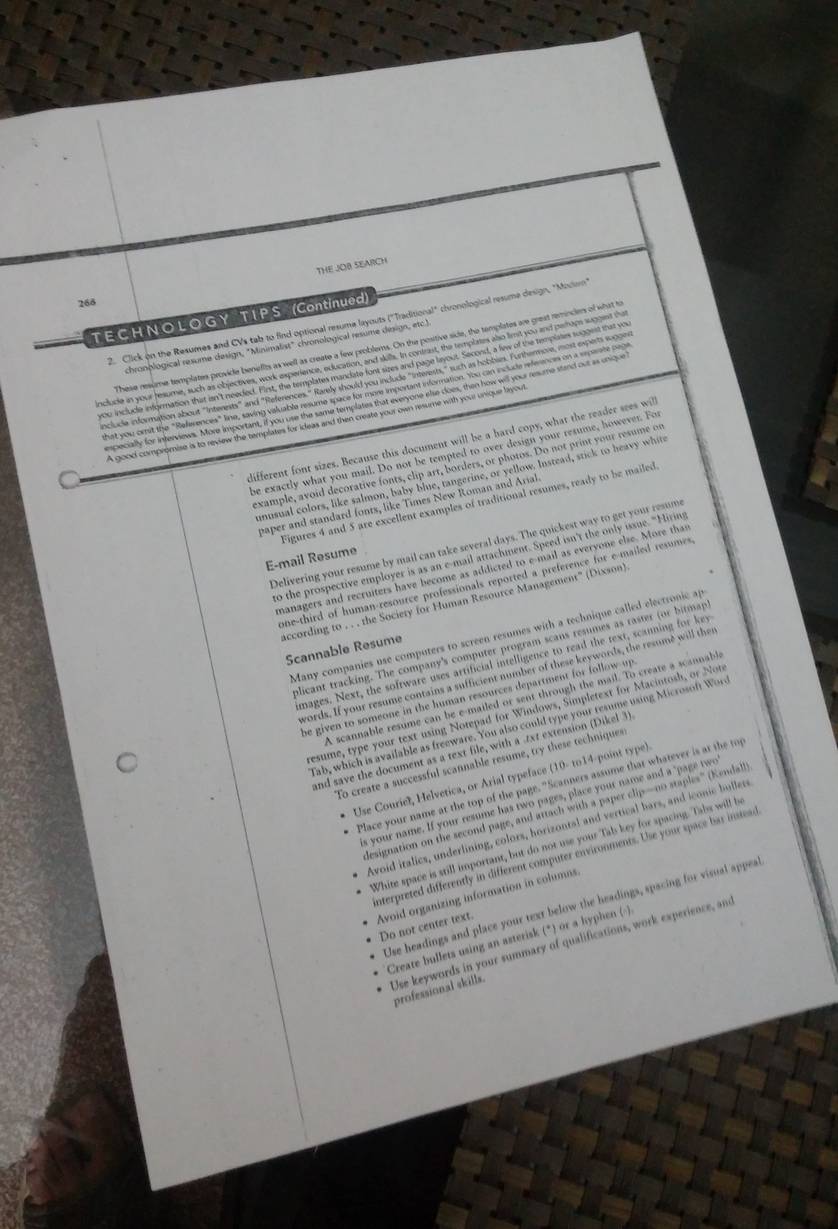}&
\includegraphics[height=4cm]{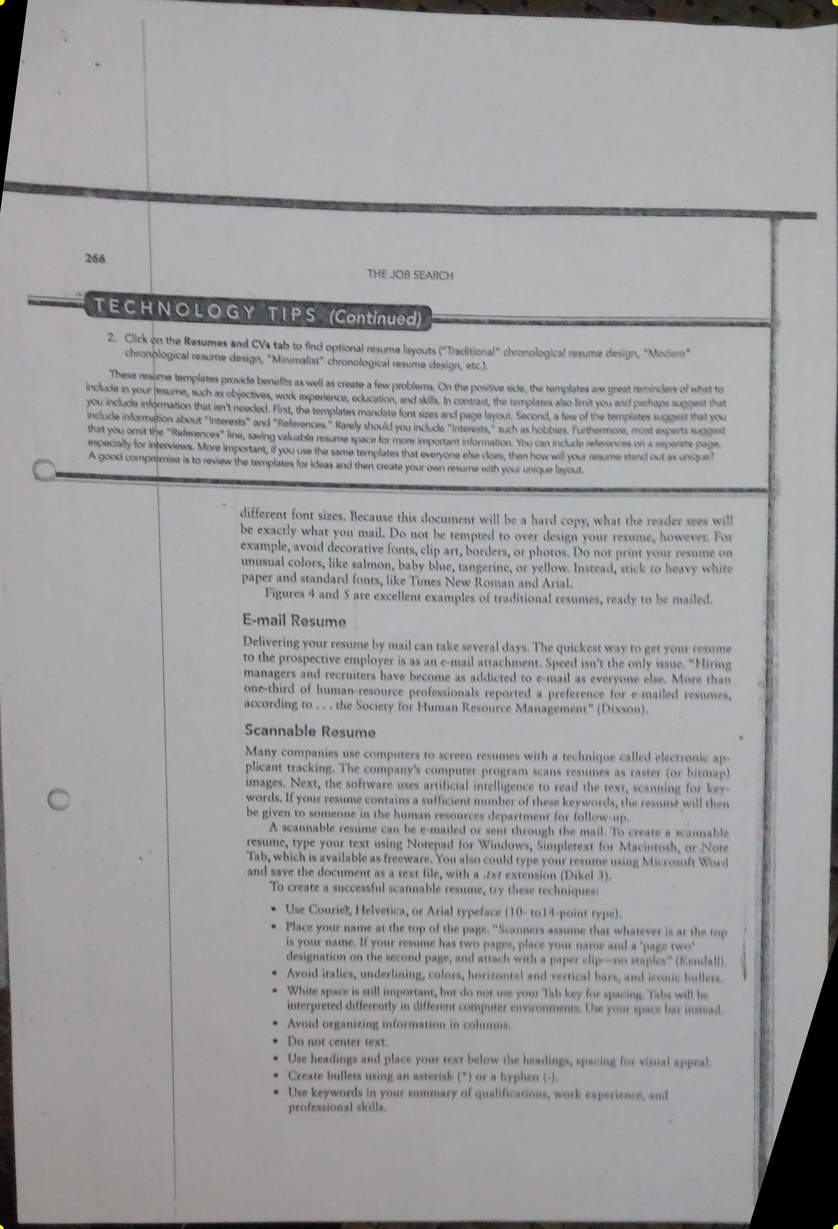} 
\end{tabular}
\caption{Results of experiment with border pixels set to zeros.} \label{figure:occluded}
\end{center}
\vspace*{-2em}
\end{figure*}
\subsection{Evaluation of Different Loss Functions}
We evaluated different loss functions (such as \loloss, \ltloss, reverse Huber) to find the best one for the problem at hand.  The \loloss loss function was able to achieve better MDE (2.59 pixels) on the validation set compared to MDE of 3.30 pixels with \ltloss loss. Actually, in all our initial experiments, \loloss performed better than \ltloss loss. This can be attributed to the fact that for the problem in hand \loloss handles extreme scenarios better than \ltloss. For instance if a corner is occluded or lies out of image frame \ltloss will give relatively high penalty and thus would be forcing the network to overfit these conditions.  We also tried the reverse Huber loss, as discussed in \cite{laina2016deeper, zwald2012berhu}, to train our convolutional neural networks. This loss function is a piecewise function of \loloss and \ltloss loss.
\begin{equation*}
 B(x)= \begin{cases} 
      \|x\| & x\leq c \\
      \|x\|^2 & x > c  \\
      \end{cases}
\end{equation*}
We validated this loss for different values of $c$ but all the results were worse than the ones obtained via \loloss loss function. In fact we found that value of $c$ is extremely sensitive to network initialization, \ie this loss function under different network initialization, with identical $c$ value produces different results. 

\subsection{RGB vs Grayscale Images}
Most of the earlier methods, discussed in section~\ref{sec:rel_work}, convert the input image to grayscale image before proceeding with perspective correction pipeline. This removes pivotal color information that can significantly help in homography estimation. For instance, color can act as an important clue for distinguishing the document from its background since majority of documents are usually white in color. 

For this experiment, we first converted training and validation set RGB images into grayscale images. Next, we trained a CNN network with same architecture as discussed in \secref{method} on the grayscale training set. The MDE obtained on the grayscale validation set was 11.59 pixels. This is far worse compared to MDE obtained using RGB images. This supports our original hypothesis that color plays an important role in recovering homography in textual documents.

\subsection{Evaluation of Synthetic Dataset Design Choices}
\label{subsec:diffdata}
For evaluating the synthetic dataset design choices, we designed a set of experiments by selecting different subsets from our final training dataset. We used the same CNN architecture for all these experiments. In our first experiment, we build a training dataset excluding the lighting variations, motion and Gaussian blurs. The model trained on this dataset gave MDE of 19.52 pixels on the validation set. From this, we can infer that having a dataset that covers large photometric variations can help in better homography estimation. Next, we created a training dataset where we only included background textures from the DTD and Brodatz datasets without including the background images from MIT indoor scenes. The network trained on this dataset gave MDE of 4.54 pixels on the validation set. This error is greater than the error obtained when indoor scenes are also added to represent background.

Our method worked well for the documents whose corners were present inside the image despite the level of noise, background clutter, lighting variation, absence of page layout clues such as text lines, \etc. However, it occasionally misfired for documents whose corners (more than one) were occluded or outside the image boundaries. To tackle this problem, we did an experiment where we set image margins pixels to zero values to simulate the occluded corners in our synthetic dataset. Precisely, we set 30 pixels from left and right image margins and 40 pixels from the top and bottom margins to zeros. However, we did not change the true annotations positions. Although, the model trained on this version of dataset was able to reliably estimate unseen corners of the documents, however, its MDE was relatively higher than the model trained on dataset without occluded corners. \figref{occluded} shows the results of our experiment on some sampled documents. We have not yet explored this any further.

Our these experiments and their results validate that design choices we made during dataset creation indeed represent the diversity of background textures, and photometric and geometric transformations the system is likely to encounter in the wild.
\subsection{Performance on SmartDoc-QA Dataset}\label{sec:sdoc}
SmartDoc-QA\footnote{\url{http://navidomass.univ-lr.fr/SmartDoc-QA/}} dataset~\cite{nayef2015smartdoc} is a recently proposed dataset for evaluating the performance of OCR systems on camera captured document images. Although it contains document images captured under different simulated conditions, these conditions are quite limited compared to wild-settings and to our proposed dataset. That is, all images in this dataset are captured: (i) across fixed red background and thus have clear contrast with background;  (ii) with only fixed set of blurs (6 different blurs) and lighting conditions (5 different). In short, this dataset lacks variations in image appearance encountered in wild settings.

\begin{figure} [!tb]
\begin{center}
\begin{tabular}{@{\hskip 2pt}c@{\hskip 2pt}c@{\hskip 2pt}c@{\hskip 2pt}c@{\hskip 2pt}c@{\hskip 2pt}}

 \includegraphics[width=0.12\textwidth]{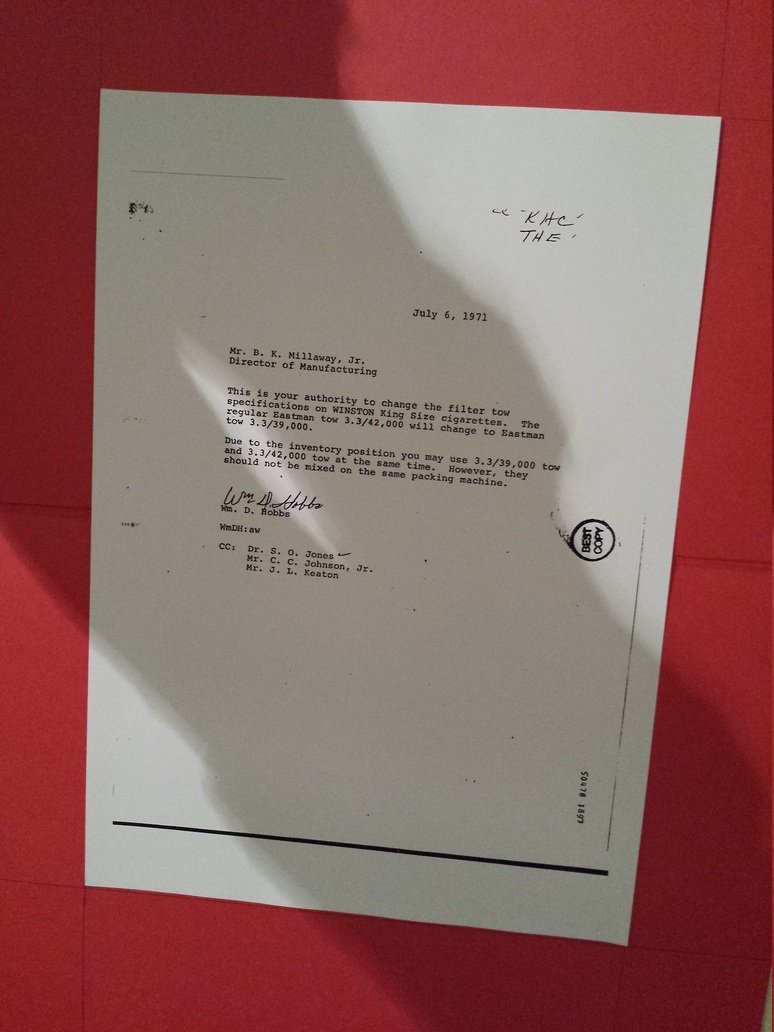} &
 \includegraphics[width=0.12\textwidth]{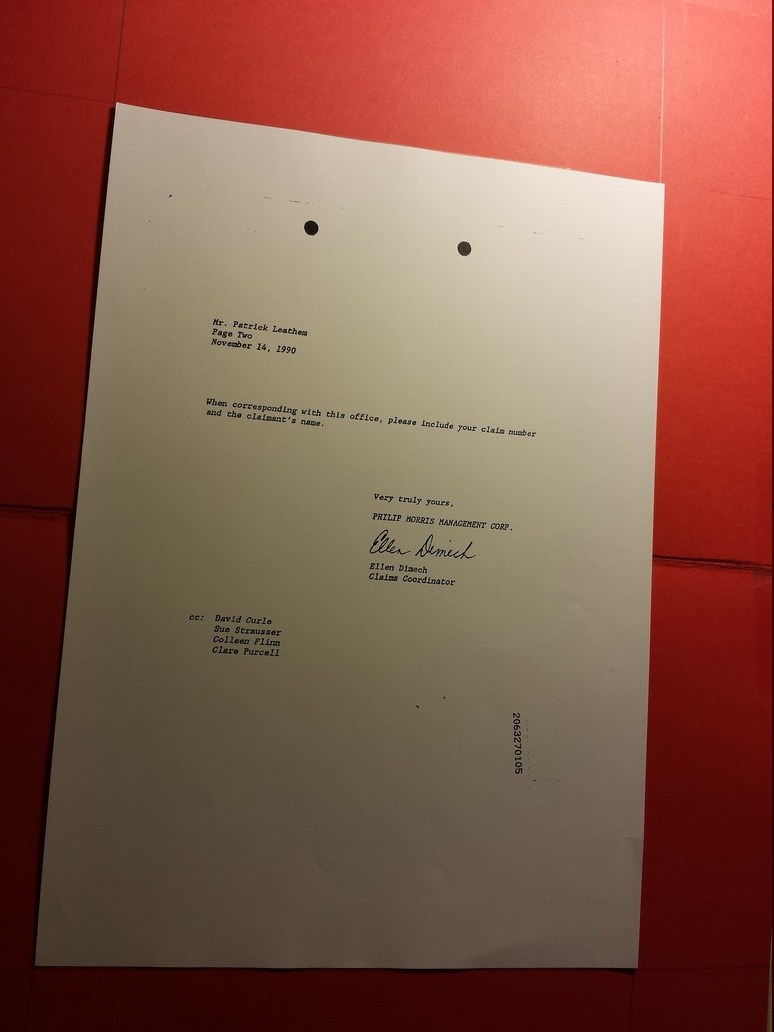} &
 \includegraphics[width=0.12\textwidth]{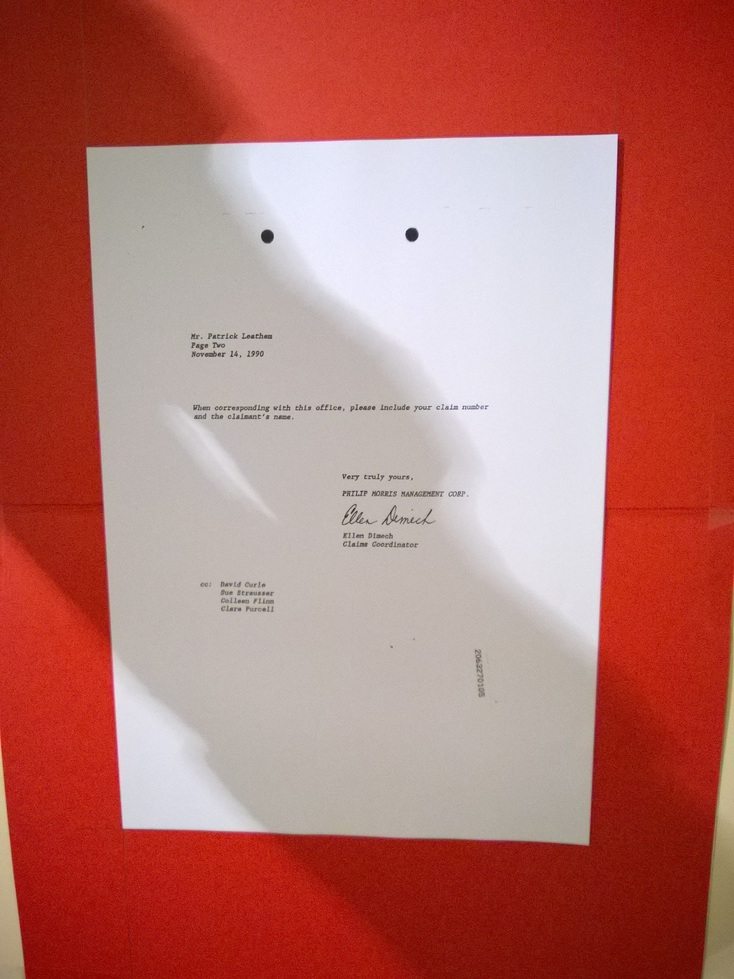} &
 \includegraphics[width=0.12\textwidth]{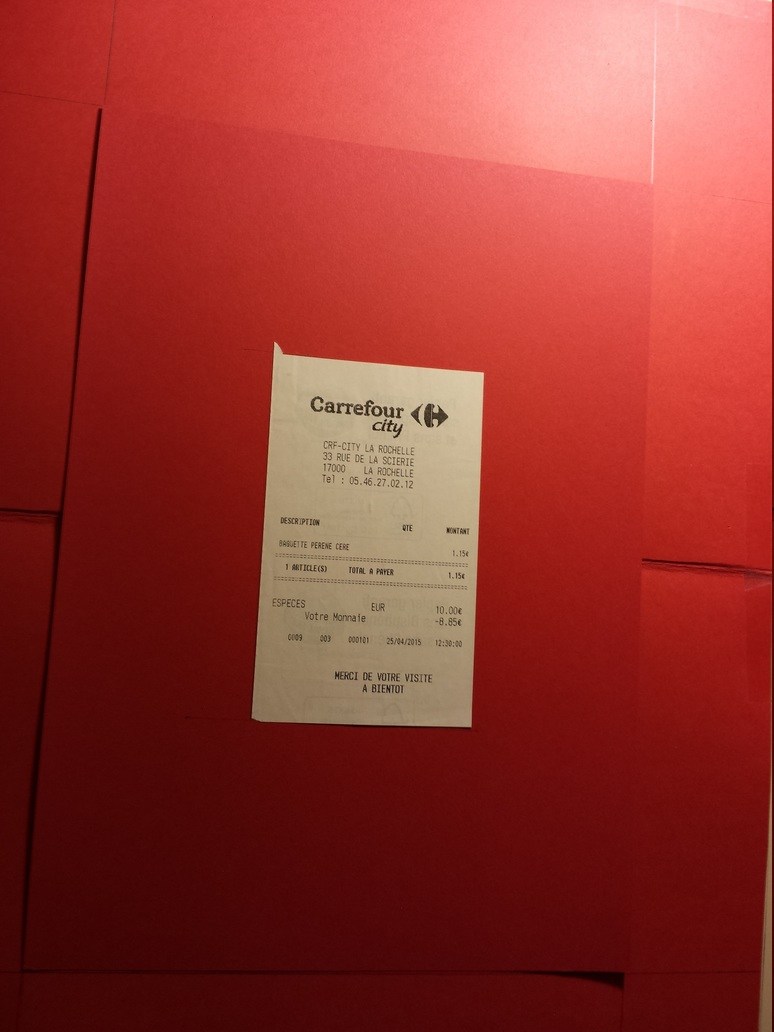}
 \\
   \includegraphics[width=0.12\textwidth]{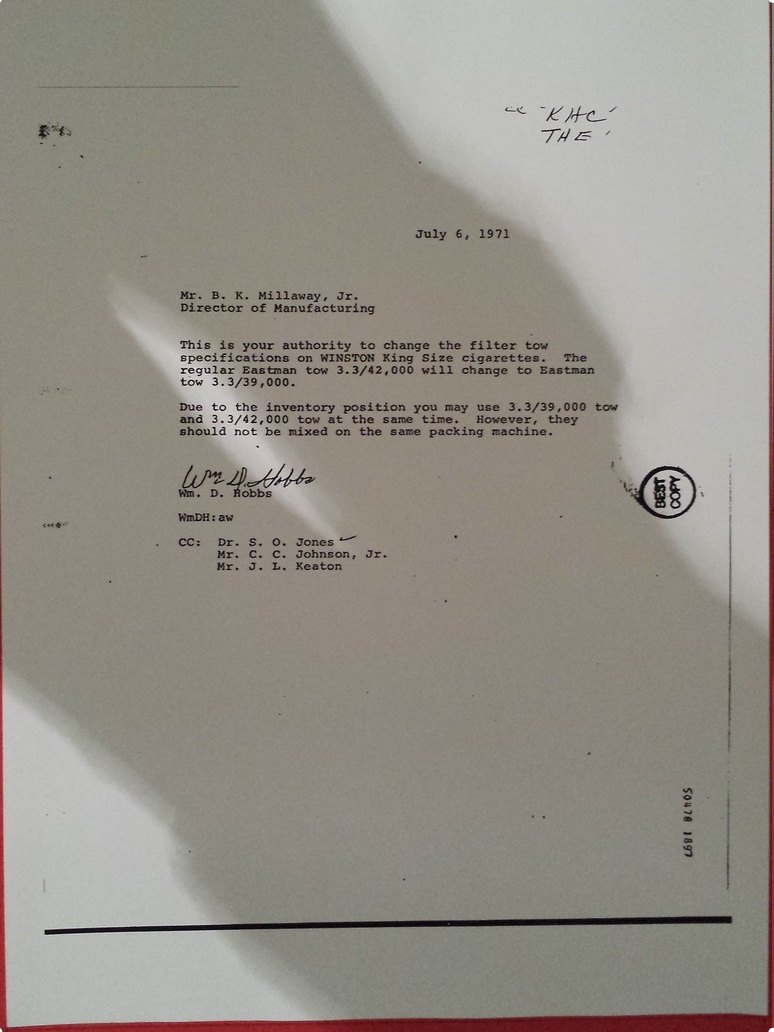} &
 \includegraphics[width=0.12\textwidth]{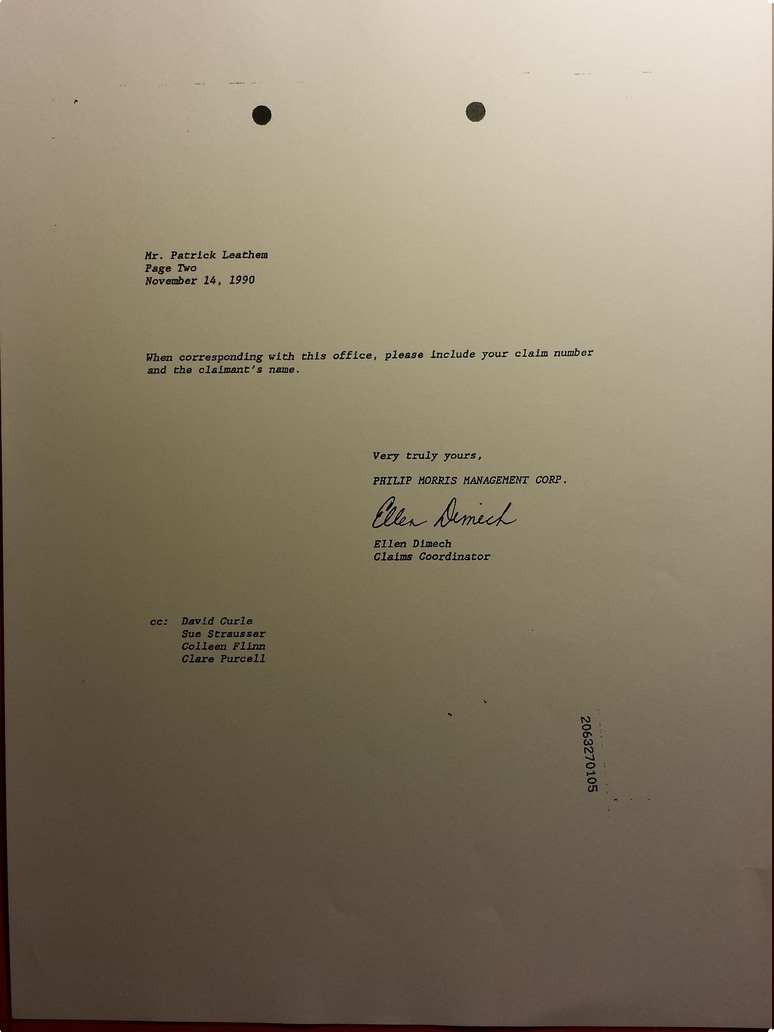} &
 \includegraphics[width=0.12\textwidth]{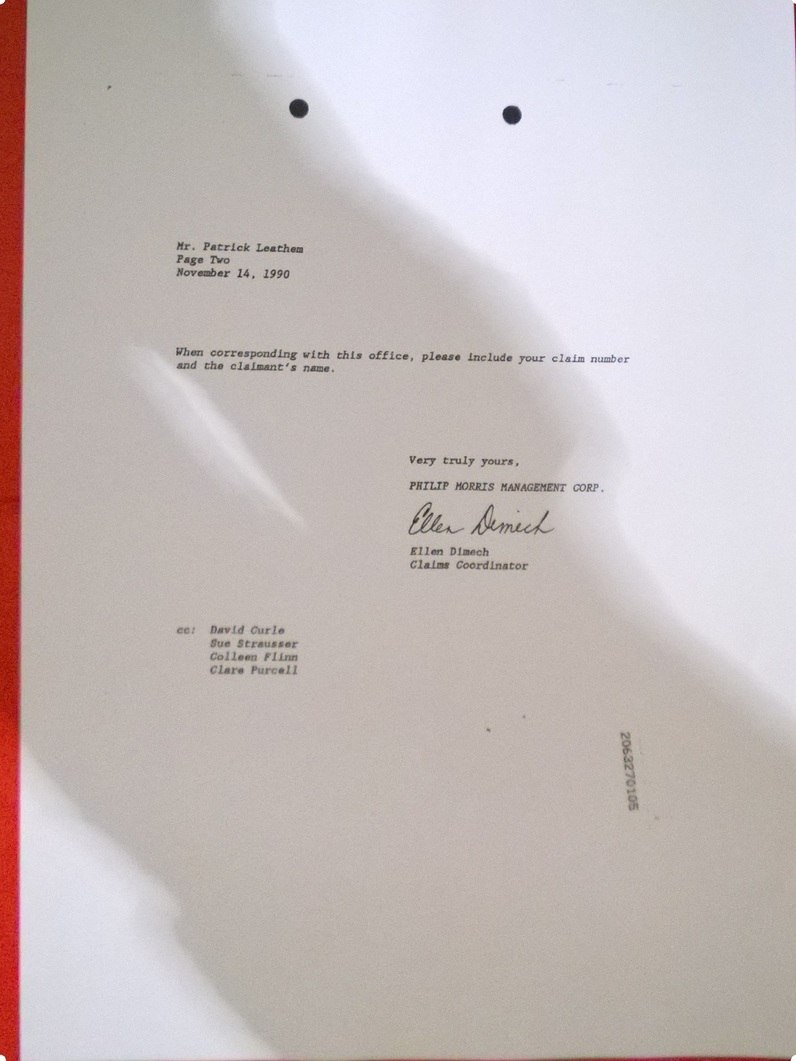} &
 \includegraphics[width=0.12\textwidth]{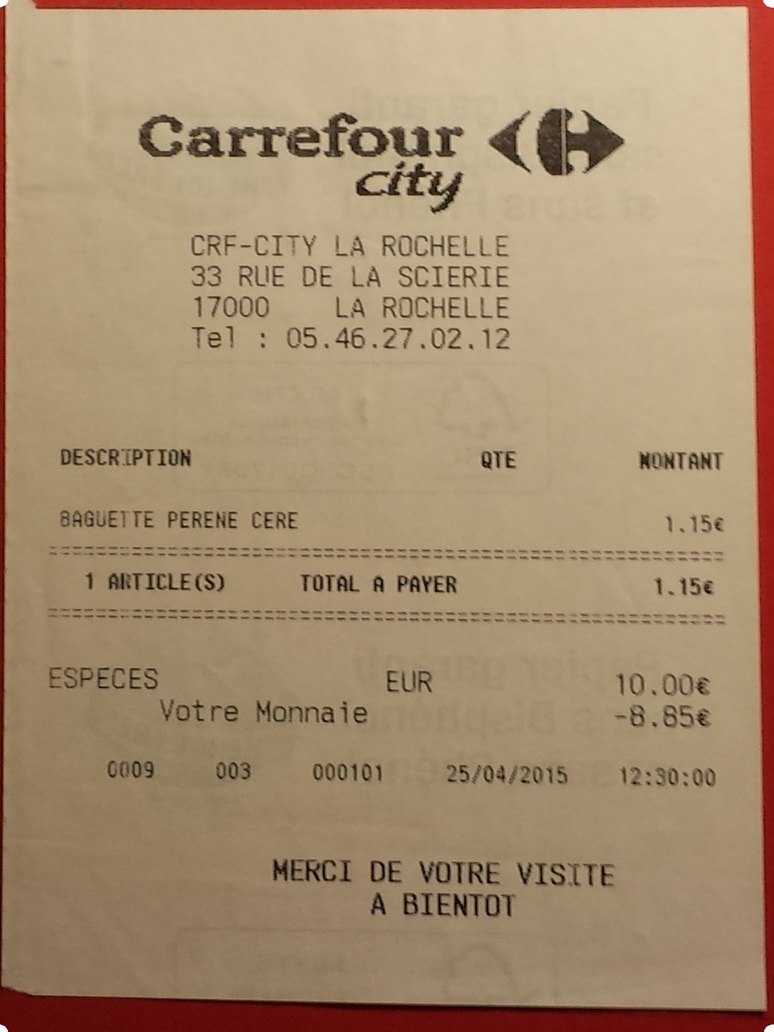}
\end{tabular}
\caption{Results of our already trained algorithm on a set of sampled images from SmartDoc-QA
dataset.}\label{figure:sd}
\end{center}
\vspace*{-2em}
\end{figure}

\figref{sd} shows the results of our algorithm on a set of sampled images from SmartDoc-QA dataset. As expected, our method is able to correctly rectify all the warped documents due to presence of strong document boundary cues.  

To throughly and analytically evaluate the effect of our system on the OCR performance, we designed another experiment where we replace the Orientation and Script Detection (OSD) module of a publicly available OCR system (Tesseract\footnote{https://github.com/Tesseract-ocr}) with ABBYY Reader and proposed perspective correction algorithms. 

\tabref{ocr-sd} shows the results of these different configurations on the SmartDoc-QA dataset. Here, we use fraction of character matches as a metric to measure the performance of OCR, \ie $\frac{2.0*M}{T}$ where $M$ is number of character matches and $T$ is total number of characters in both documents.  Our algorithm improves the performance of Tesseract OCR over default OSD system as well as give on-par performance to ABBYY reader. Note that here the difference is all because of superior performance of proposed image rectification algorithm. Furthermore, Tesseract OCR bad performance is due to presence of significant motion-blur at the character level in SmartDoc-QA dataset which is leading to failure of character recognition pipeline.

\begin{table*}[!htbp]
\begin{minipage}{0.5\linewidth}
\begin{center}
\resizebox{\linewidth}{!}{
\begin{tabular}{|l|c|c|}
\hline
Methods & SmartDoc Dataset \\
\hline
OSD + Tesseract OCR & 11.61\% \\ 
ABBYY + Tesseract OCR & 16.18\% \\ 
\textbf{Our Method + Tesseract OCR} & 16.14\% \\ 
\hline
\end{tabular}
 } \\
\caption{Performance of Tesseract OCR with different perspective correction modules on SmartDoc-QA dataset.}
\end{center}\label{table:ocr-sd}
\end{minipage}
\begin{minipage}{0.5\linewidth}
\begin{center}
\begin{tabular}{|l|c|c|}
\hline
 \multirow{2}{*}{Methods}& \multicolumn{2}{c|}{Dataset}\\
 \cline{2-3}
 &\textbf{Simple} & \textbf{Complex}  \\
 
\hline
OSD + Tesseract OCR & 10.99\% & 2.94\% \\ 
ABBYY + Tesseract OCR & 23.78\% & 10.04\% \\ 
CamScanner + Tesseract OCR & 30.64\% & 7.55\% \\ 
\textbf{Our Method + Tesseract OCR} & \textbf{31.58}\% & \textbf{14.35}\%\\ 
\hline
\end{tabular}
\caption{OCR performance for different image rectification algorithms on  Simple and Complex variants of datasets from SmartDoc-QA.} \label{table:ocr-sdoc-syn}
\end{center}
\end{minipage}
\vspace*{-2em}
\end{table*}
\begin{figure*} [!t]
\begin{center}
\begin{tabular}{@{\hskip 2pt}c@{\hskip 2pt}c@{\hskip 2pt}c@{\hskip 2pt}c@{\hskip 2pt}}
\bf{Original} & \bf{Our Method} & \bf{CamScanner} & \bf{ABBYY FineReader}\\
\includegraphics[height=5cm]{images/pic10.jpg}&
\includegraphics[height=5cm]{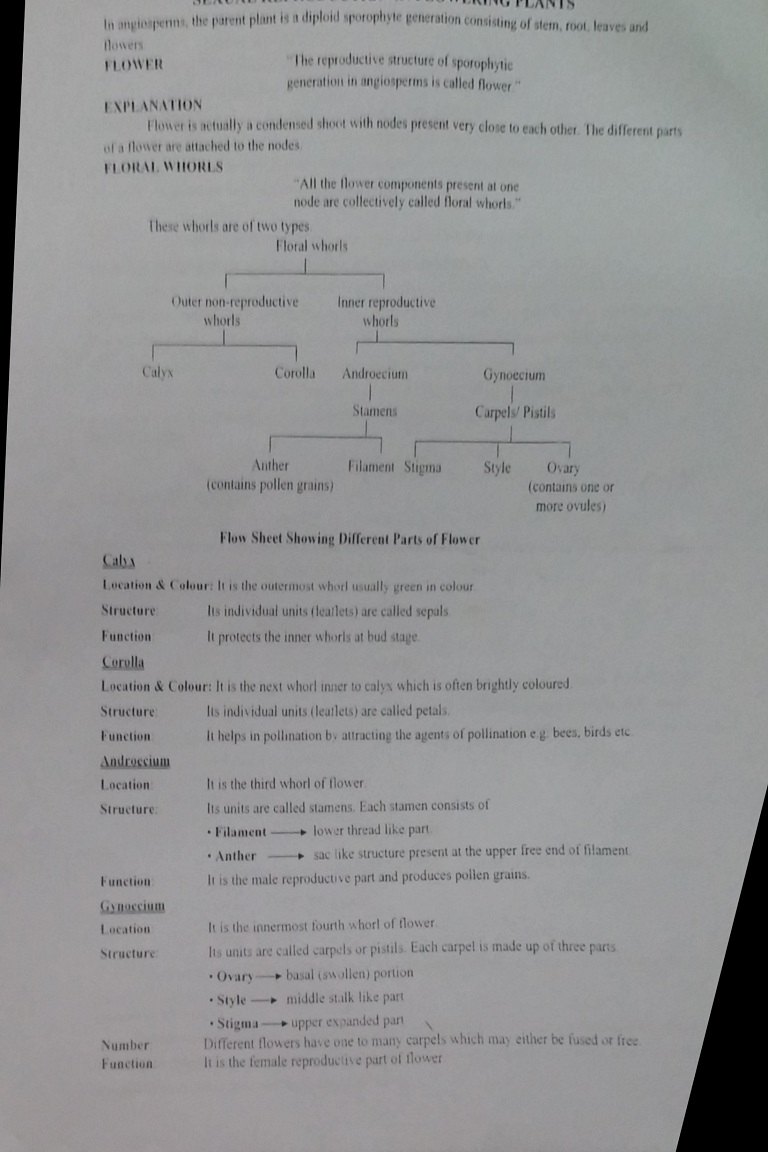}&
\includegraphics[height=5cm]{images/pic10.jpg} &
\includegraphics[height=5cm]{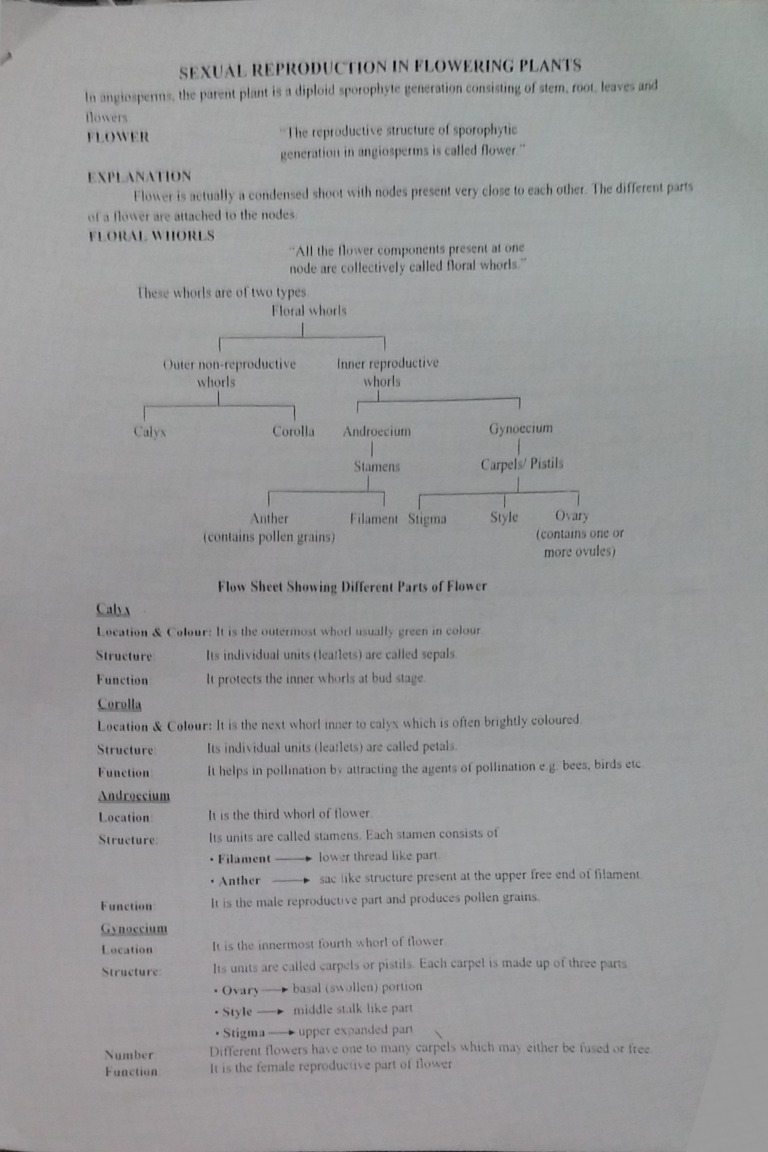} \\
\includegraphics[height=5cm]{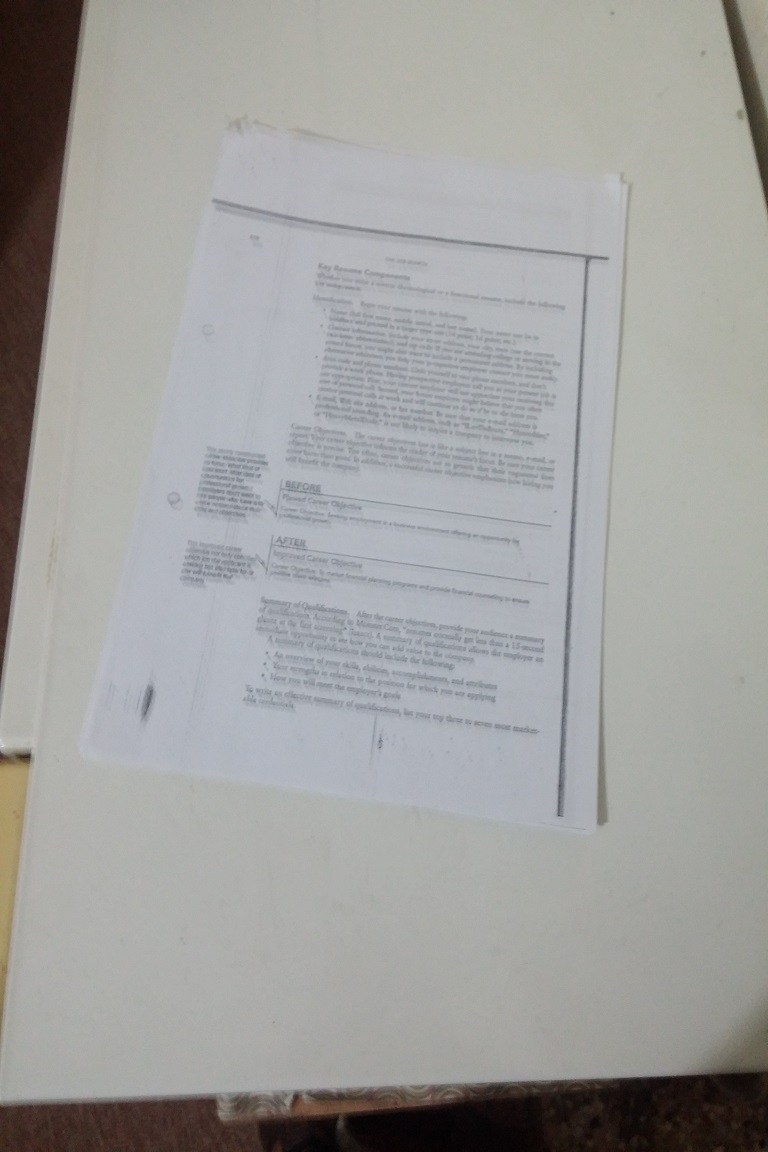}&
\includegraphics[height=5cm]{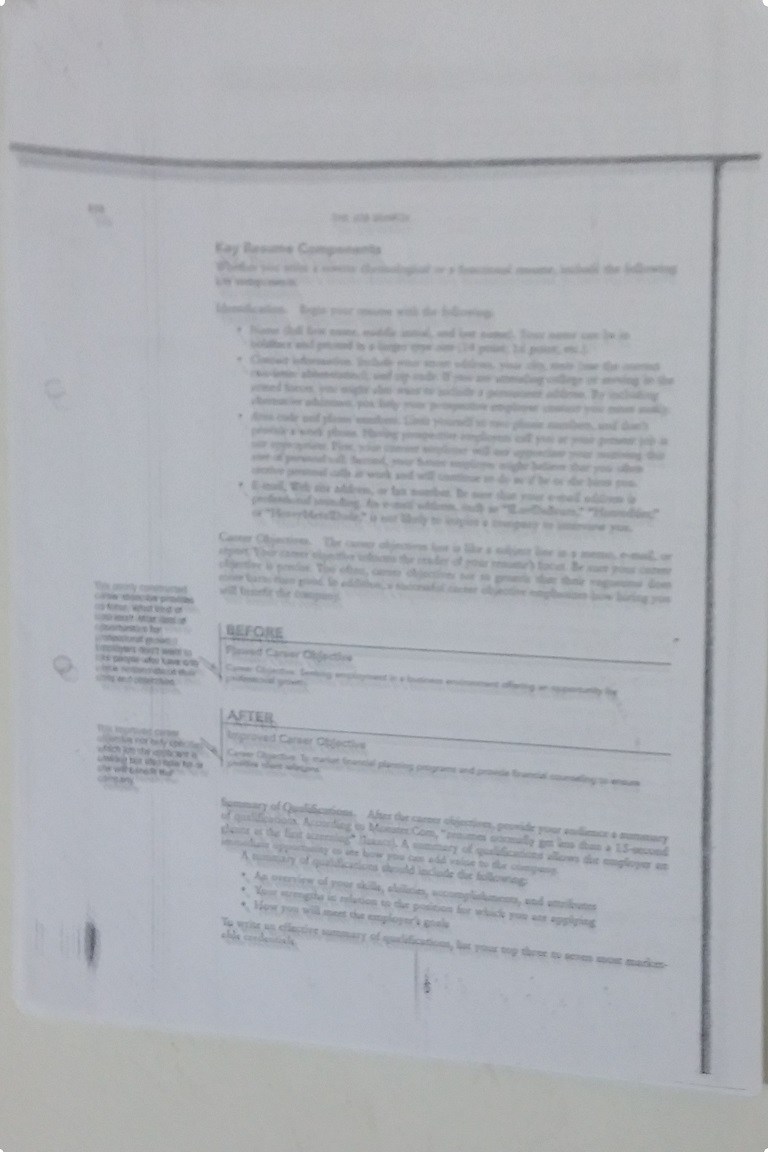} &
\includegraphics[height=5cm]{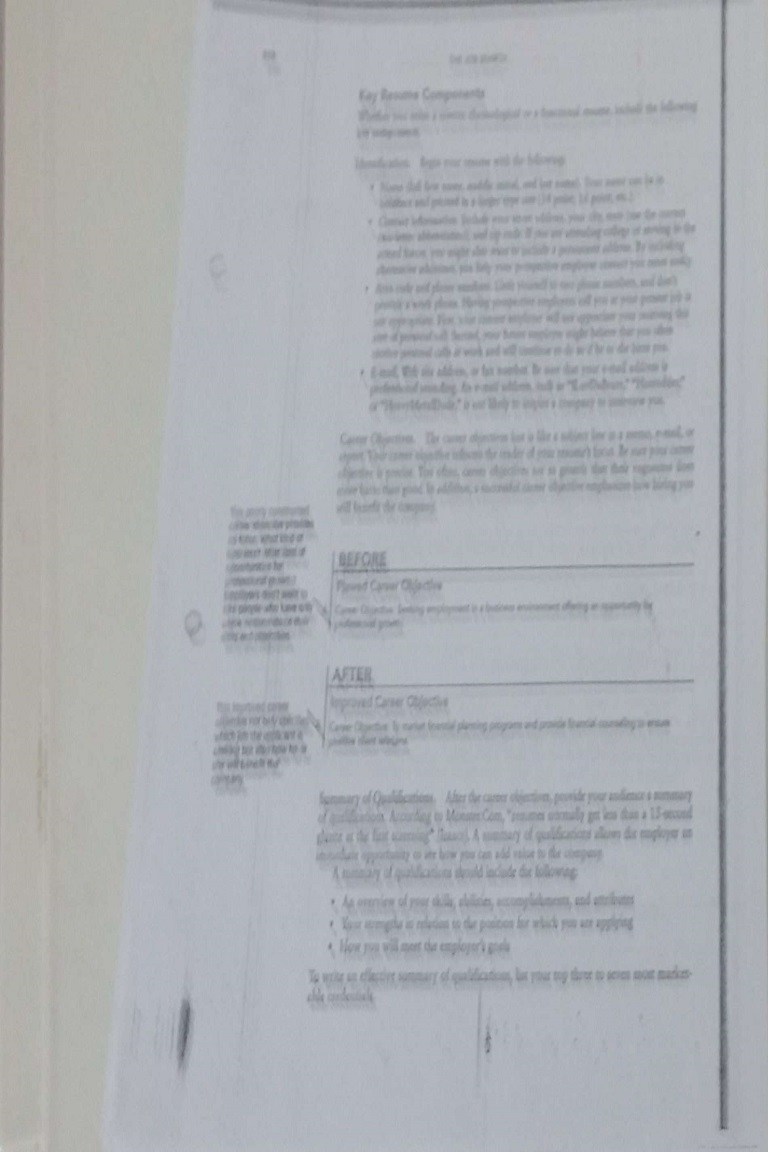} &
\includegraphics[height=5cm]{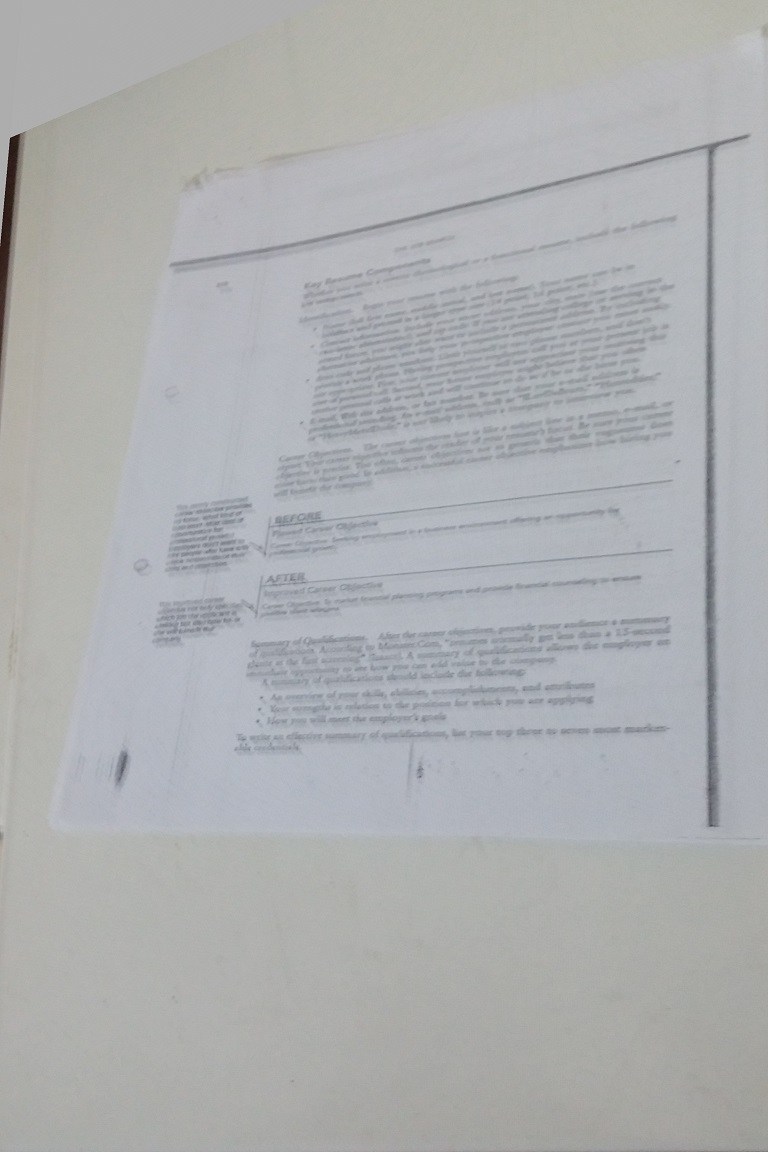} \\
\includegraphics[height=5cm]{images/different_categories/normal_results/pic9.jpg} &
\includegraphics[height=5cm]{images/different_categories/normal_results/wpic9.jpg} &
\includegraphics[height=5cm]{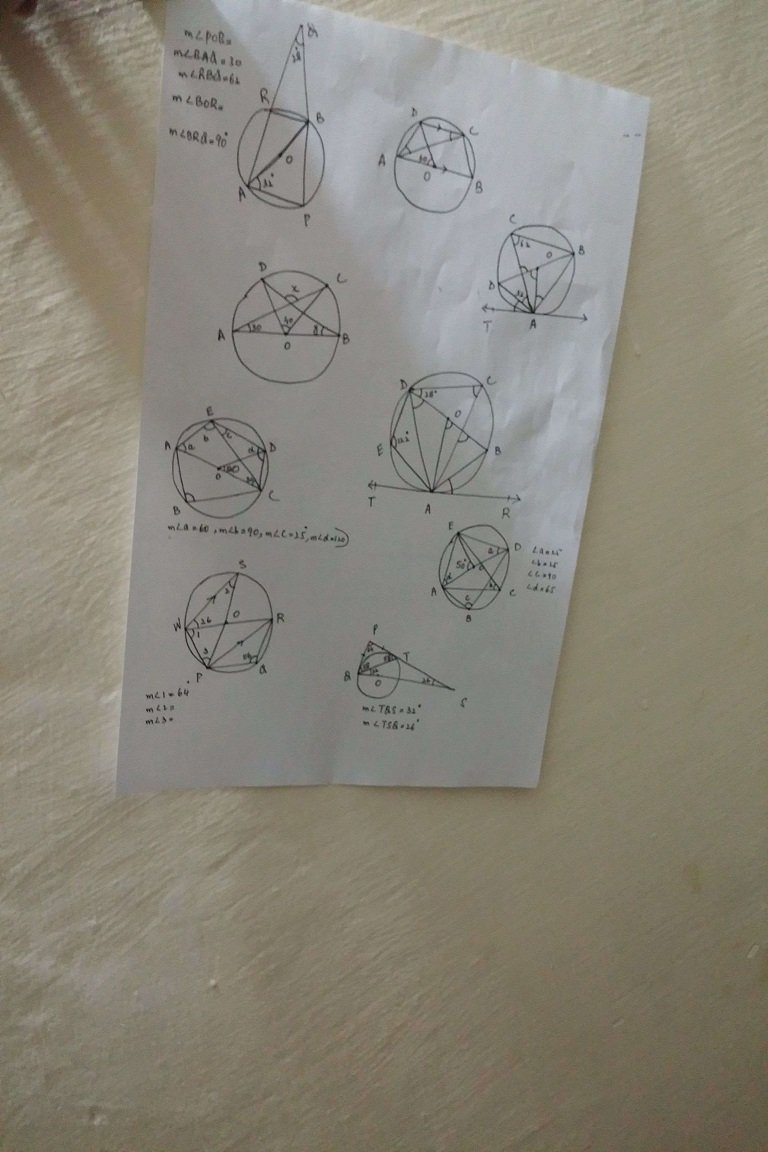} &
\includegraphics[height=5cm]{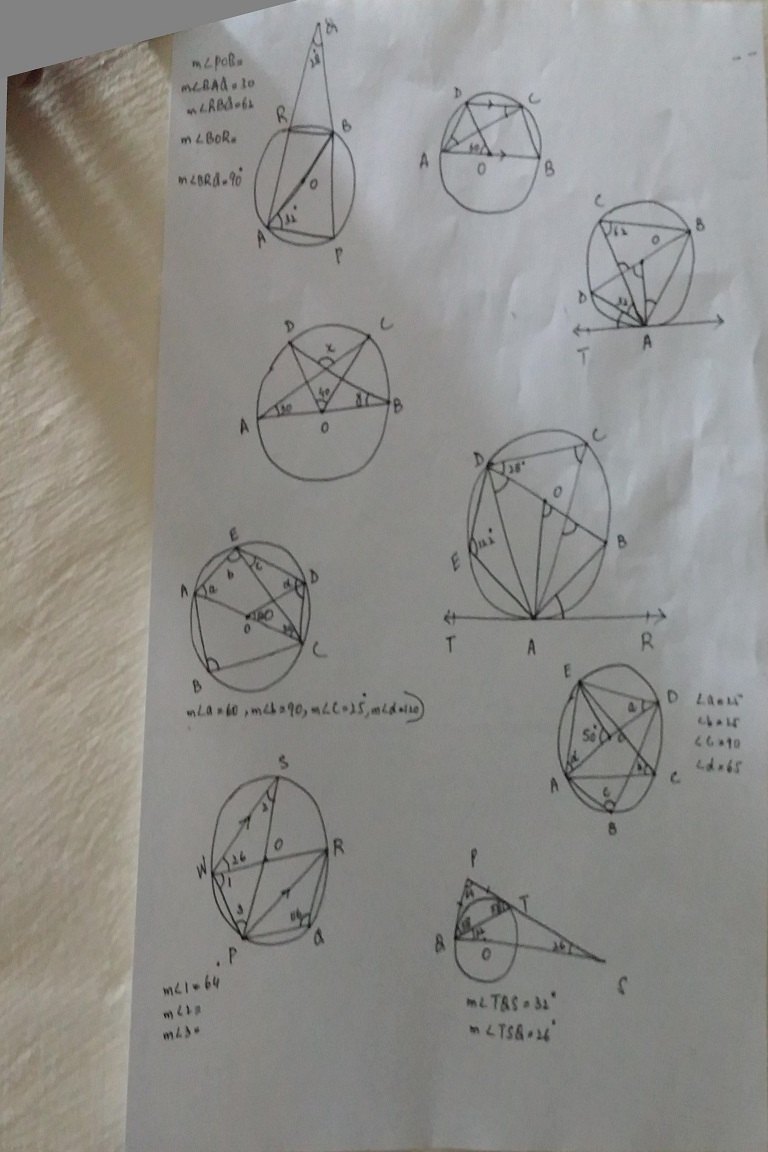} \\
\includegraphics[height=5cm]{images/different_categories/normal_results/pic15.jpg} &
\includegraphics[height=5cm]{images/different_categories/normal_results/wpic15.jpg} &
\includegraphics[height=5cm]{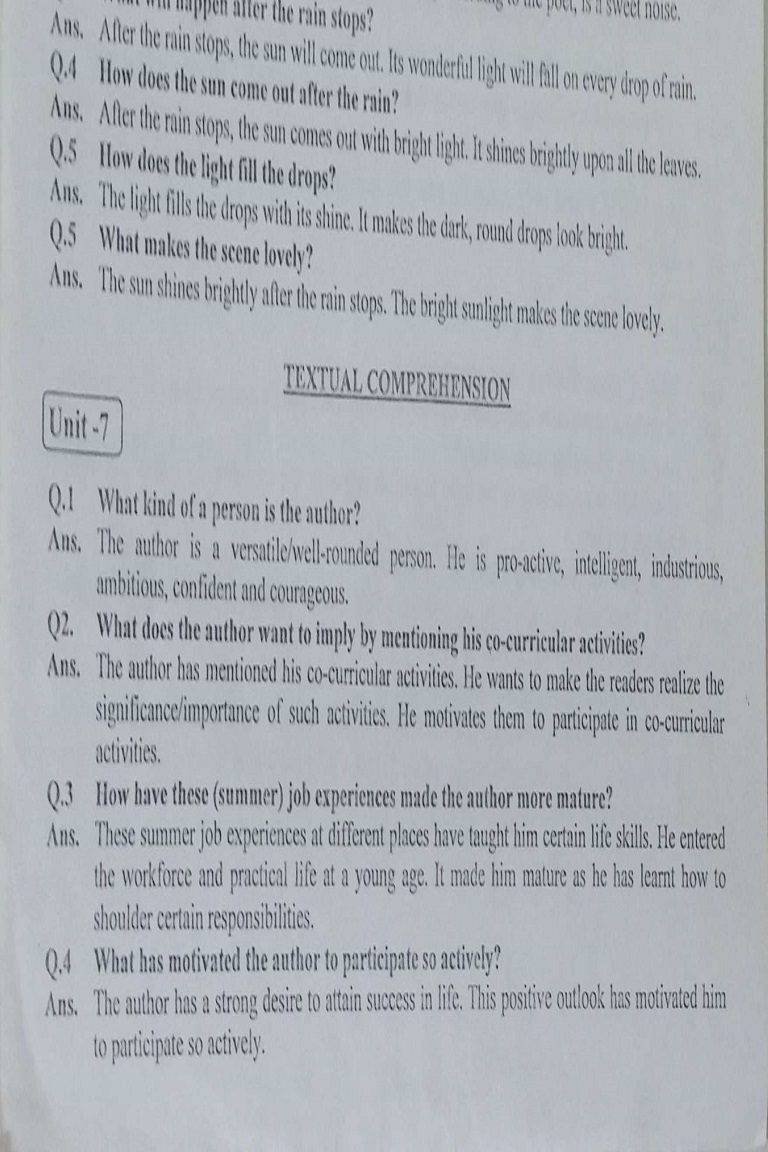} &
\includegraphics[height=5cm]{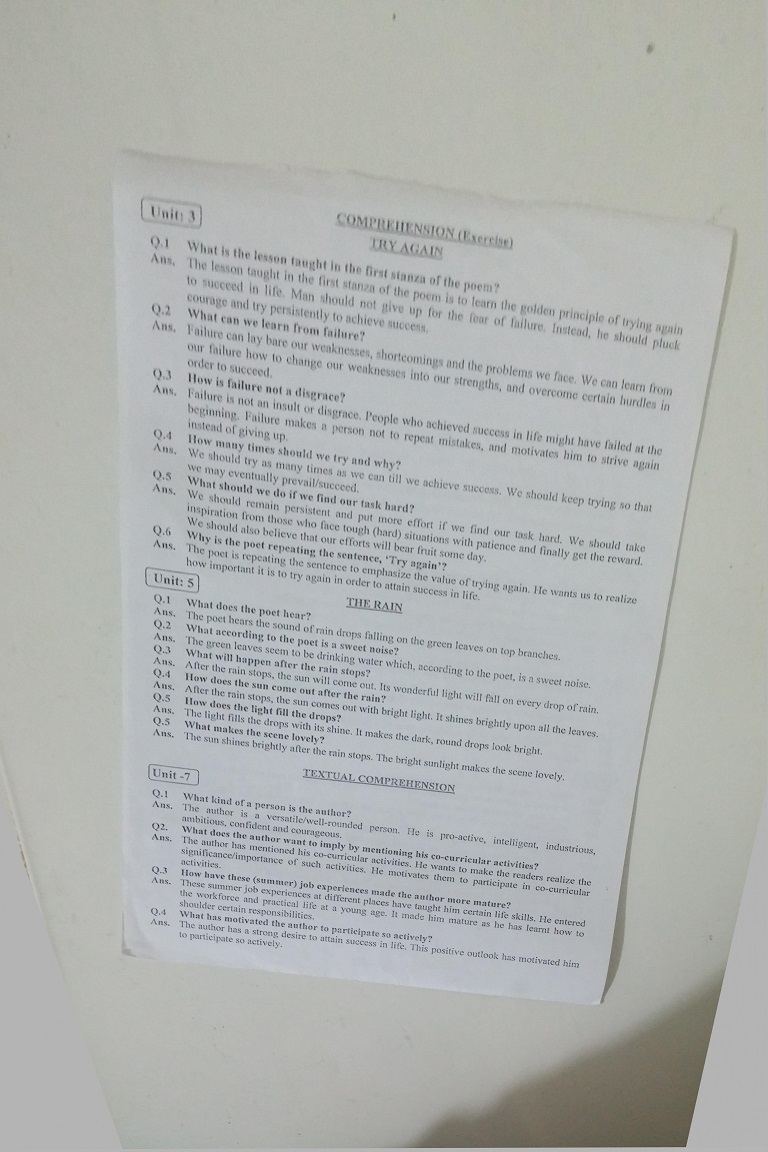} 
\end{tabular}
\caption{Visual comparison of results produced by our method and popular commercial softwares. On this sample set, our method is outperforming both the other softwares.}
\label{figure:software_comp}
\end{center}
\vspace*{-2em}
\end{figure*}
\begin{figure*} [!th]
\begin{center}
\begin{tabular}{cccc}
\bf{Original} & \bf{Our Method} & \bf{CamScanner} & \bf{ABBYY FineReader}\\
\includegraphics[width=0.20\textwidth]{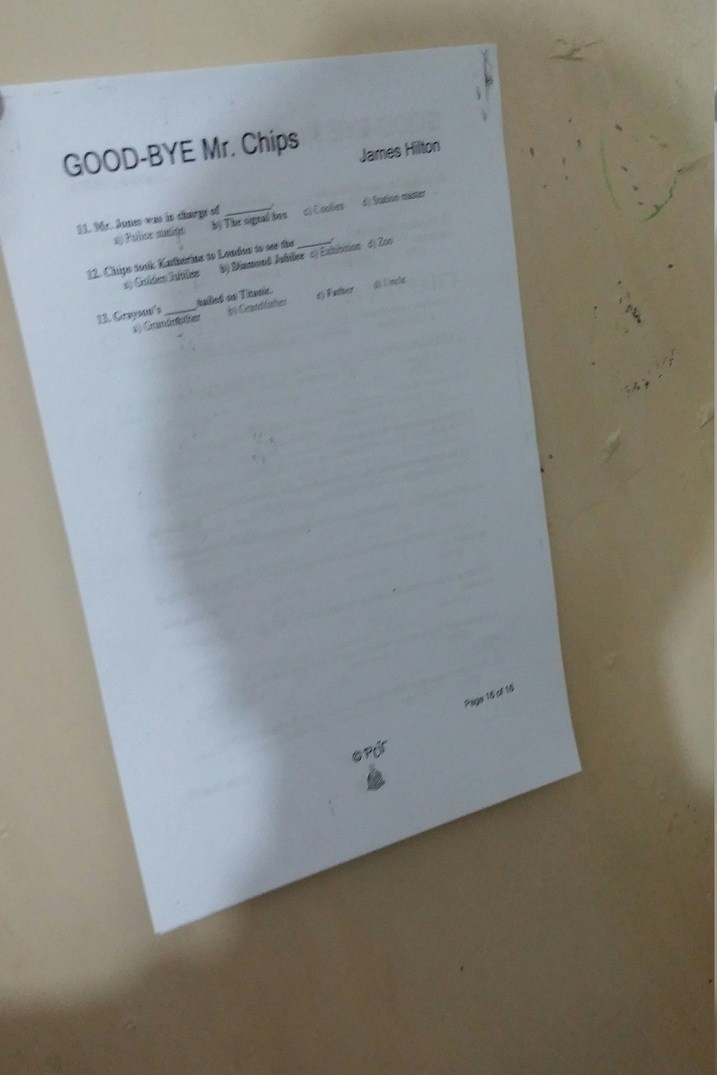} &
\includegraphics[width=0.20\textwidth]{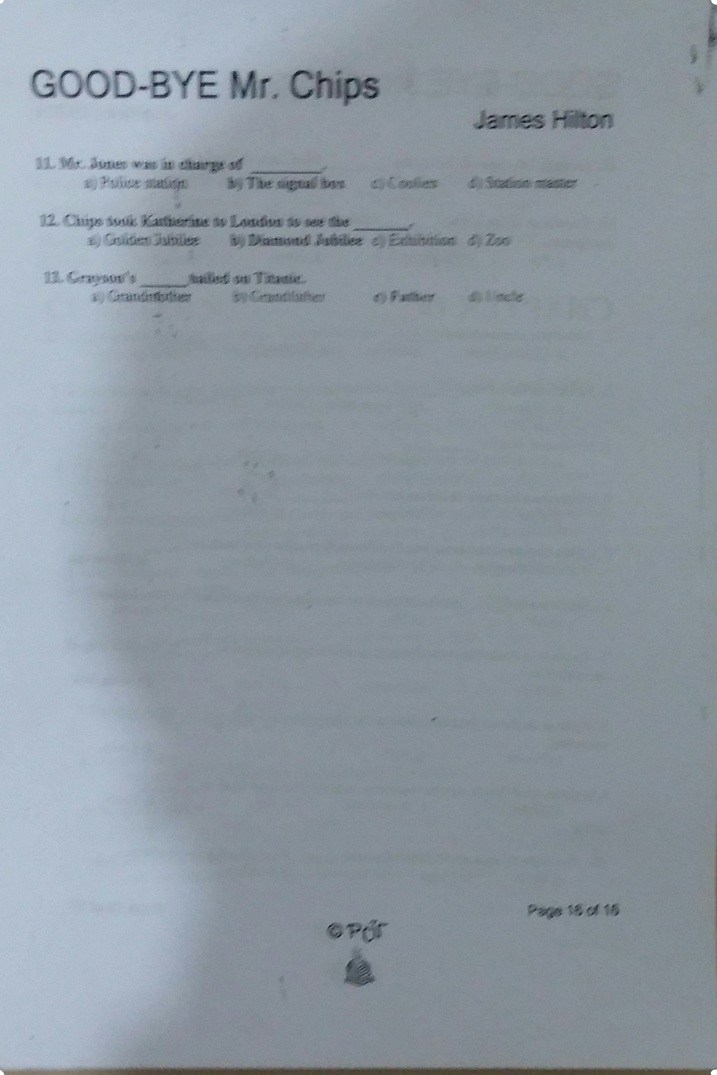} &
\includegraphics[width=0.20\textwidth]{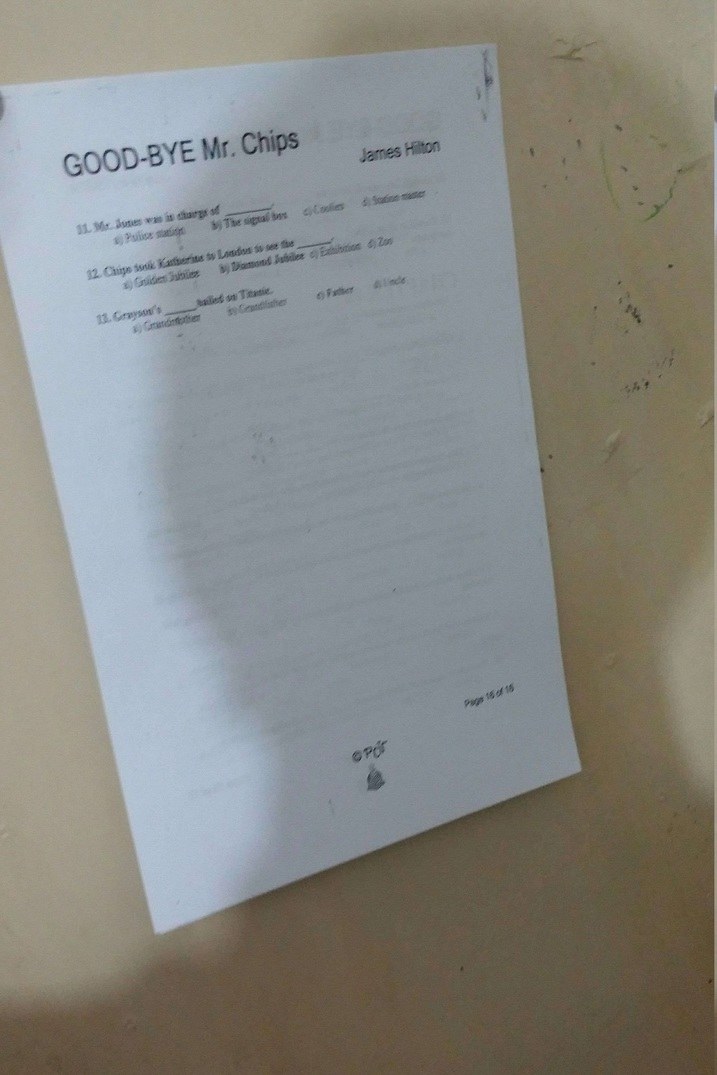} &
\includegraphics[width=0.20\textwidth]{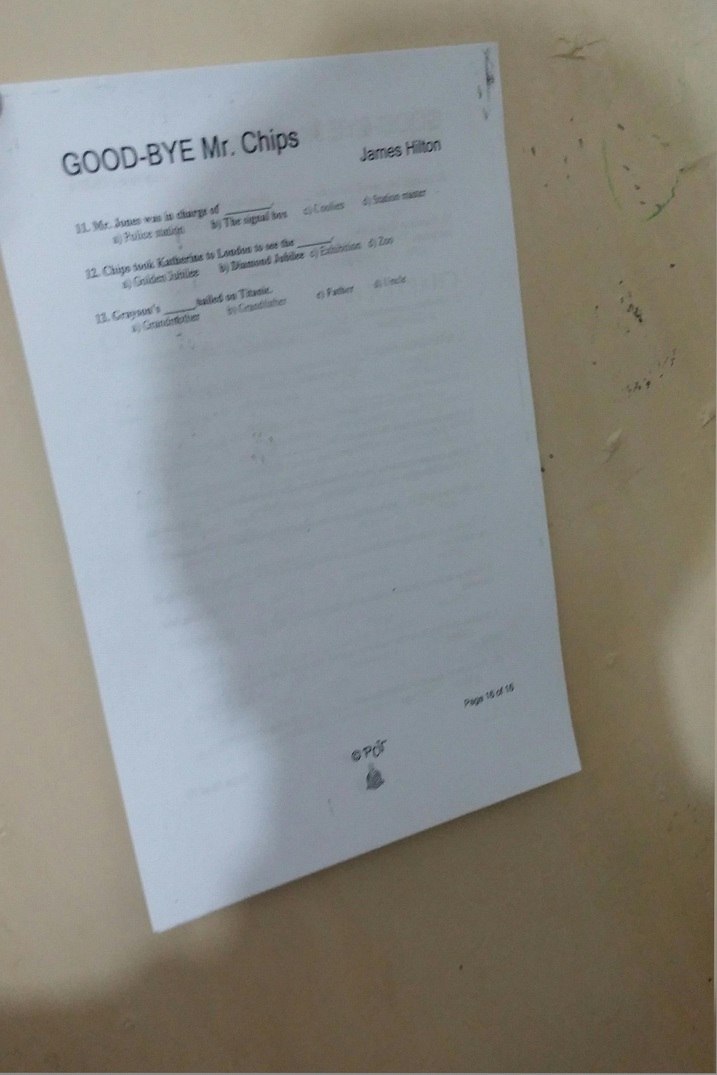} \\
\includegraphics[width=0.20\textwidth]{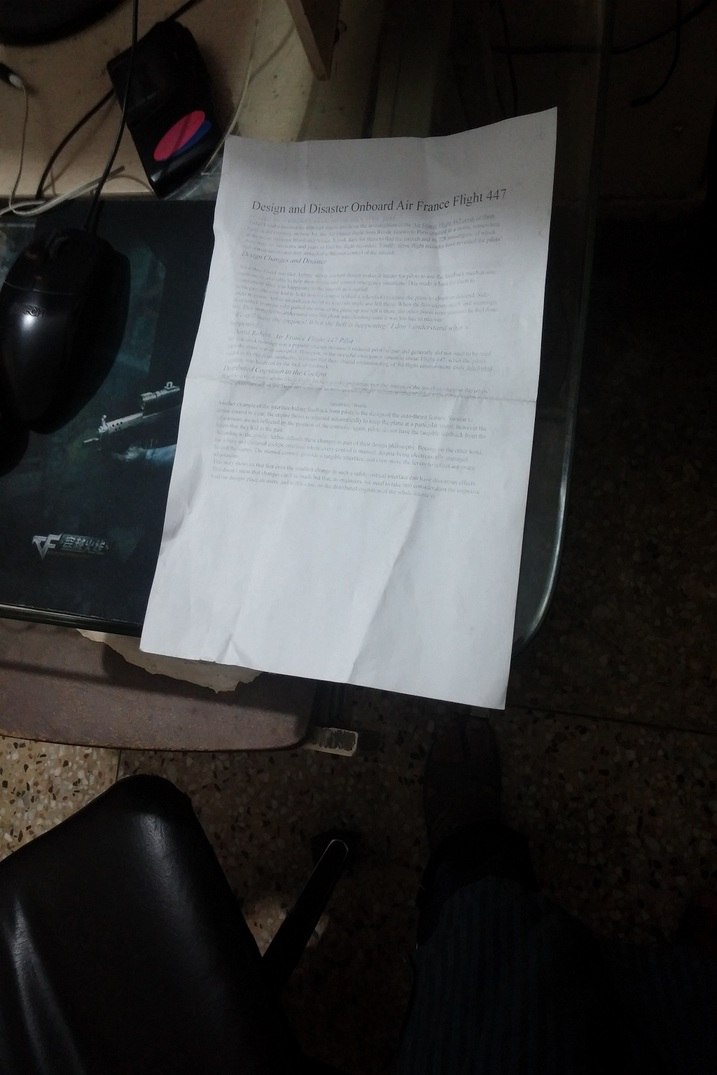} &
\includegraphics[width=0.20\textwidth]{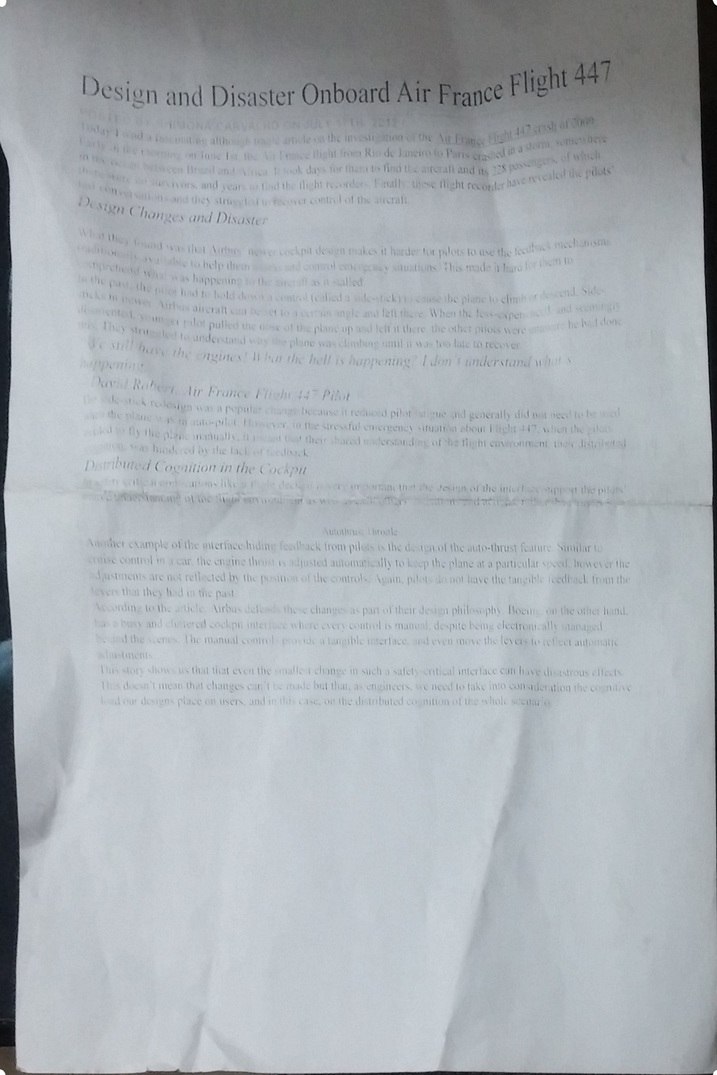} &
\includegraphics[width=0.20\textwidth]{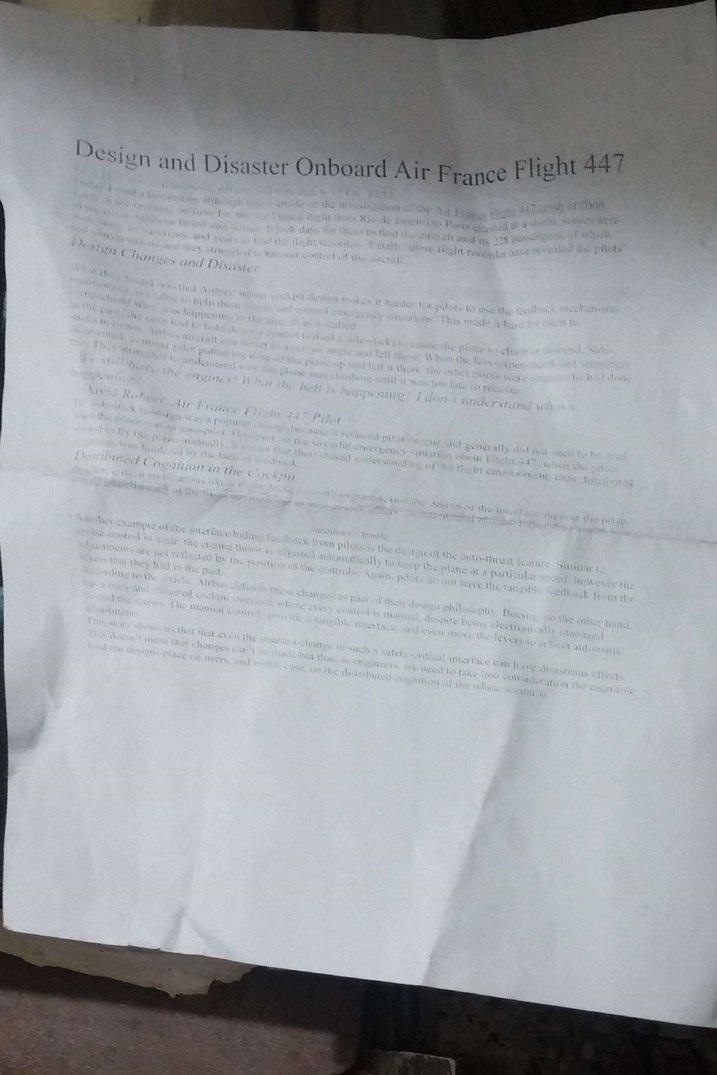} &
\includegraphics[width=0.20\textwidth]{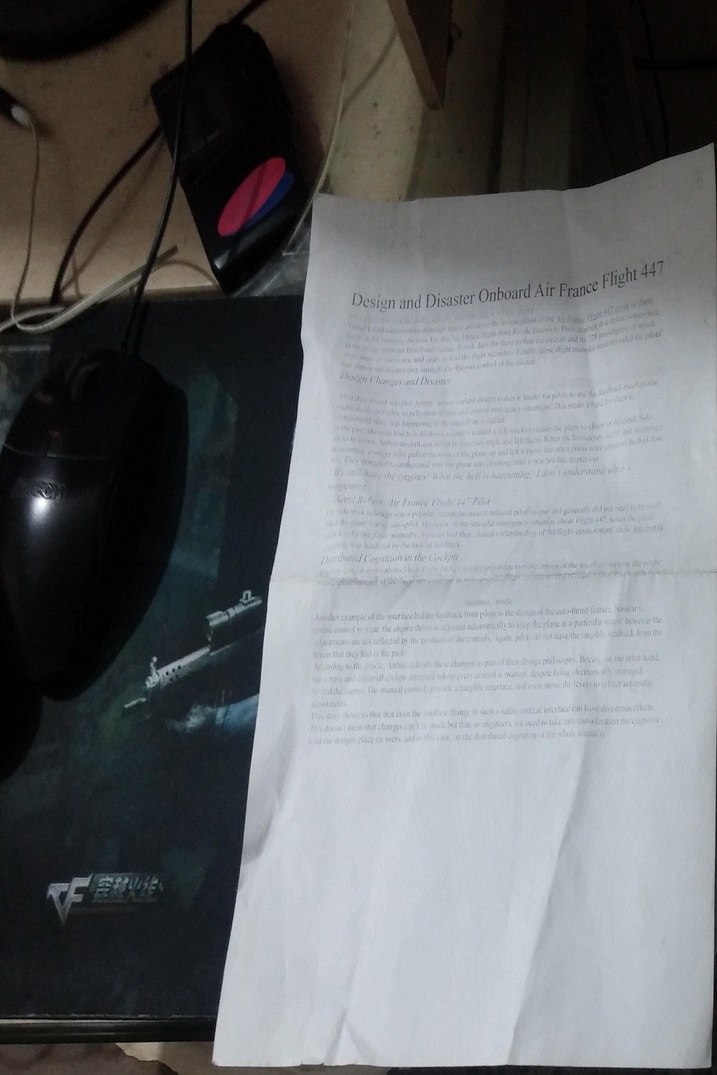} \\
\includegraphics[width=0.20\textwidth]{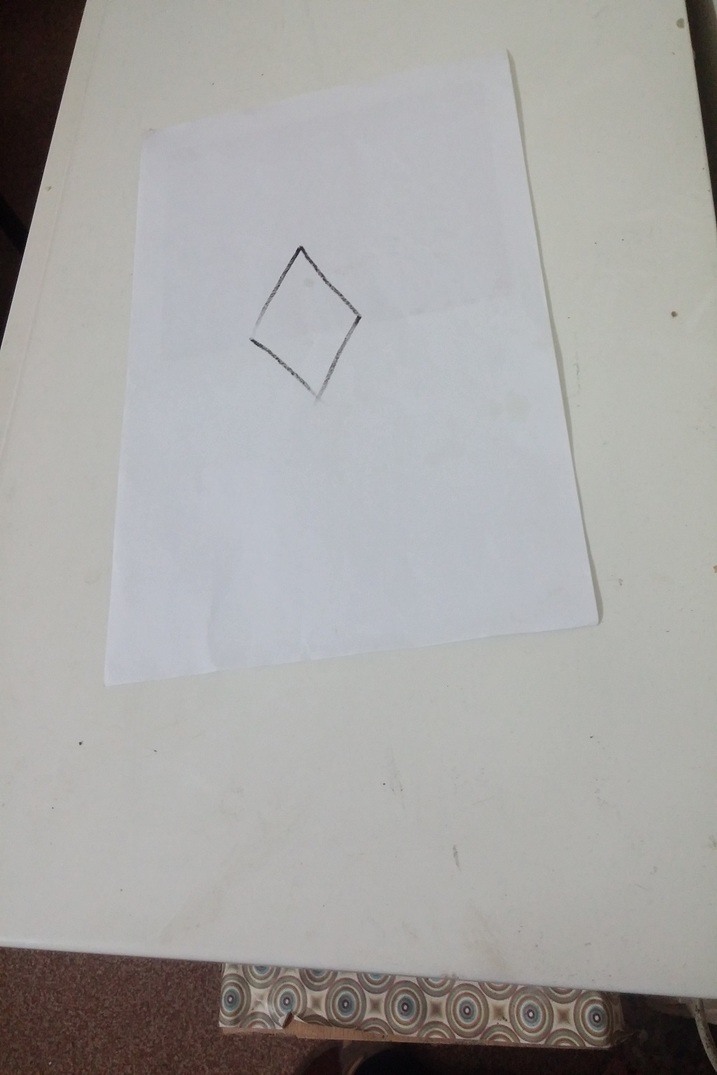} &
\includegraphics[width=0.20\textwidth]{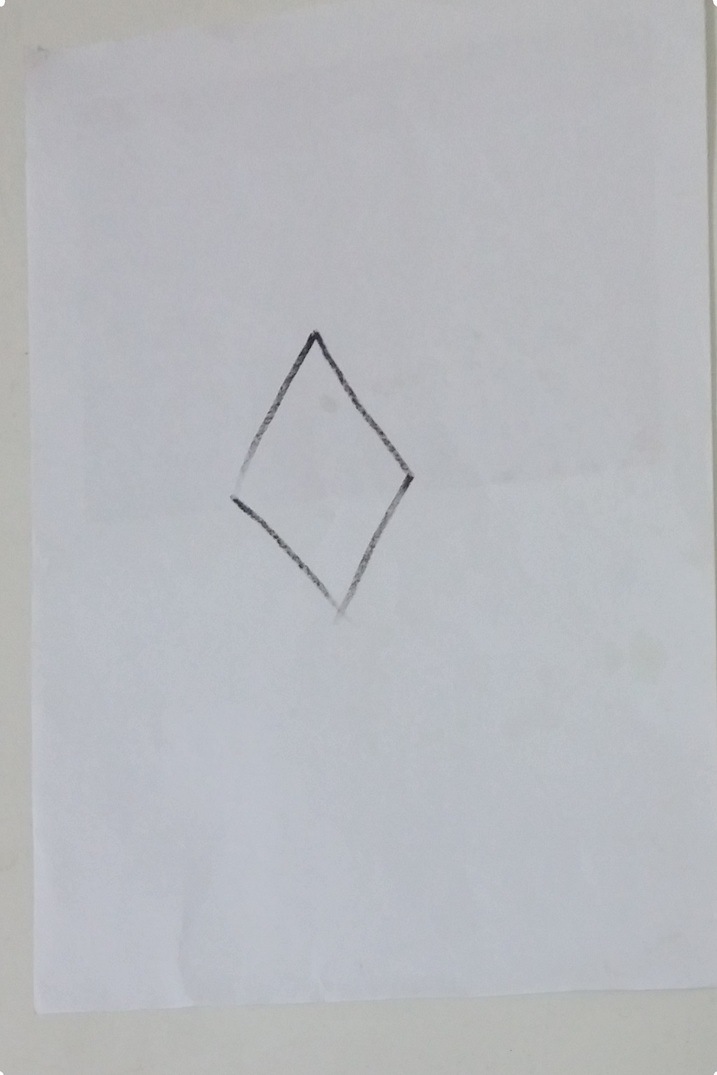} &
\includegraphics[width=0.20\textwidth]{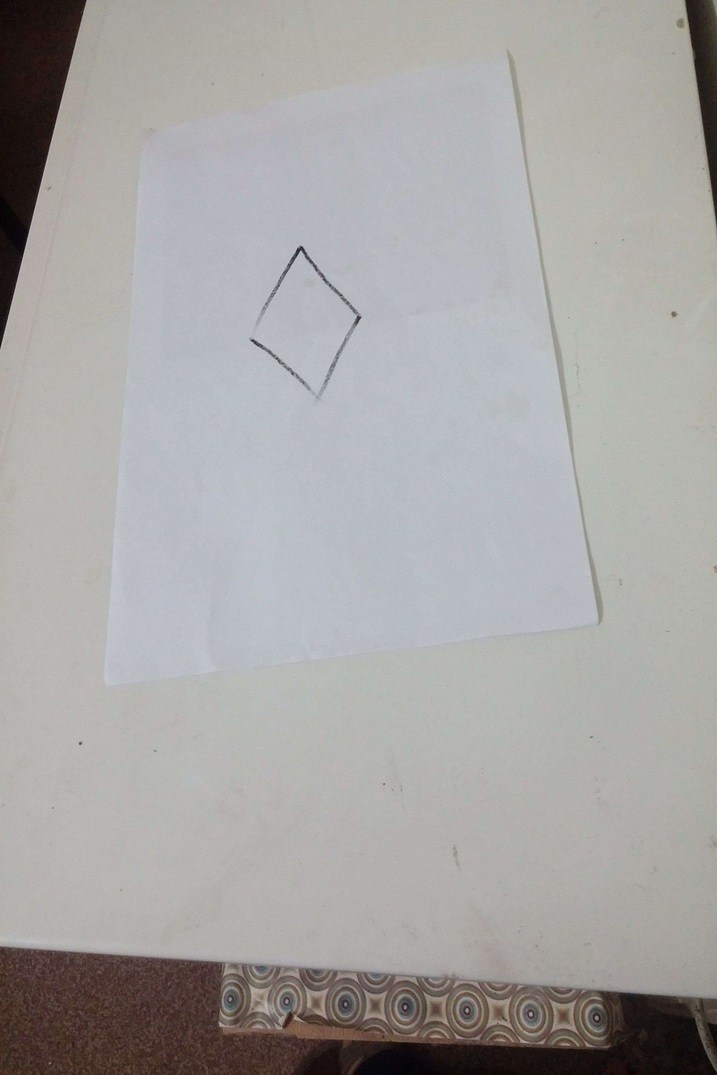} &
\includegraphics[width=0.20\textwidth]{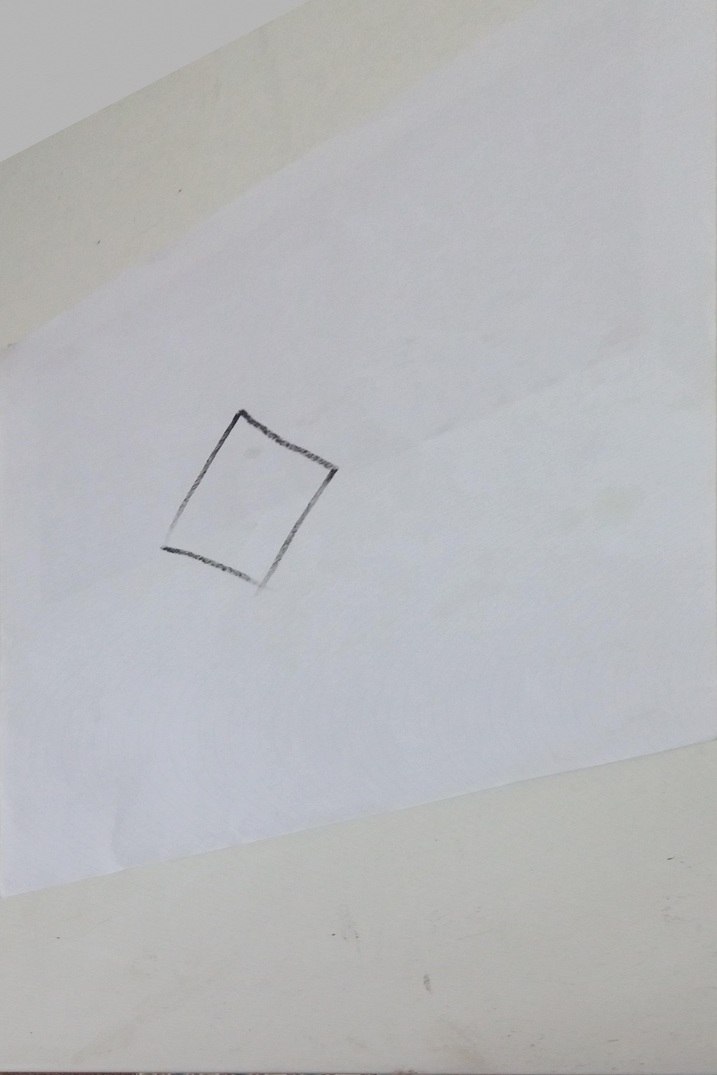} \\
\includegraphics[width=0.20\textwidth]{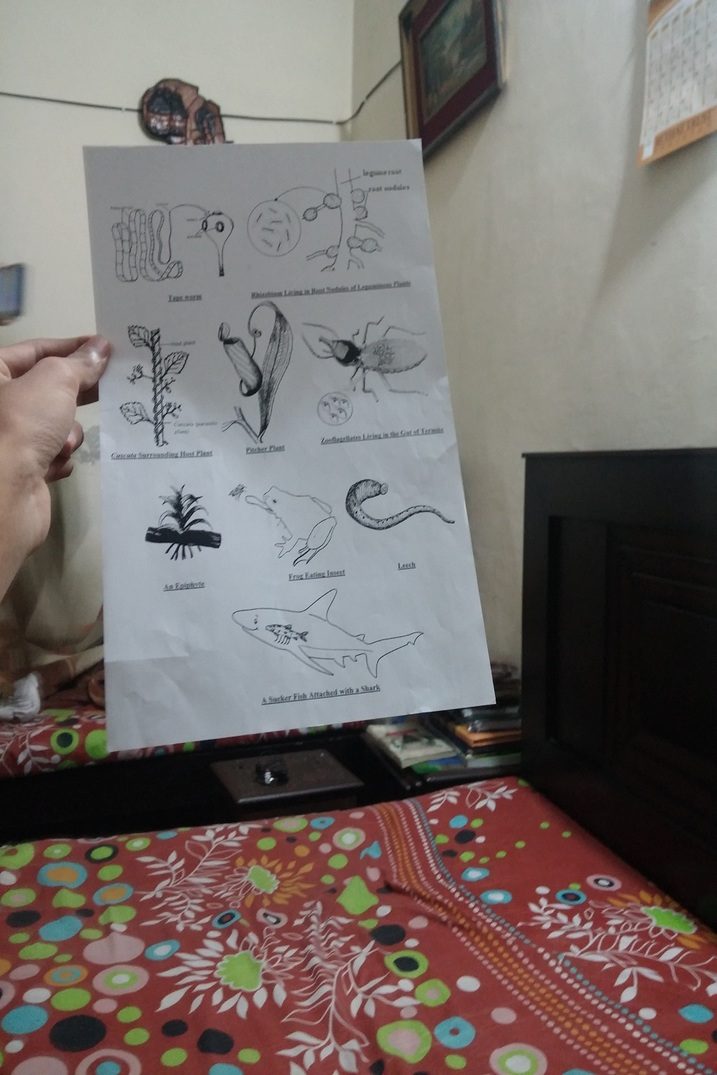} &
\includegraphics[width=0.20\textwidth]{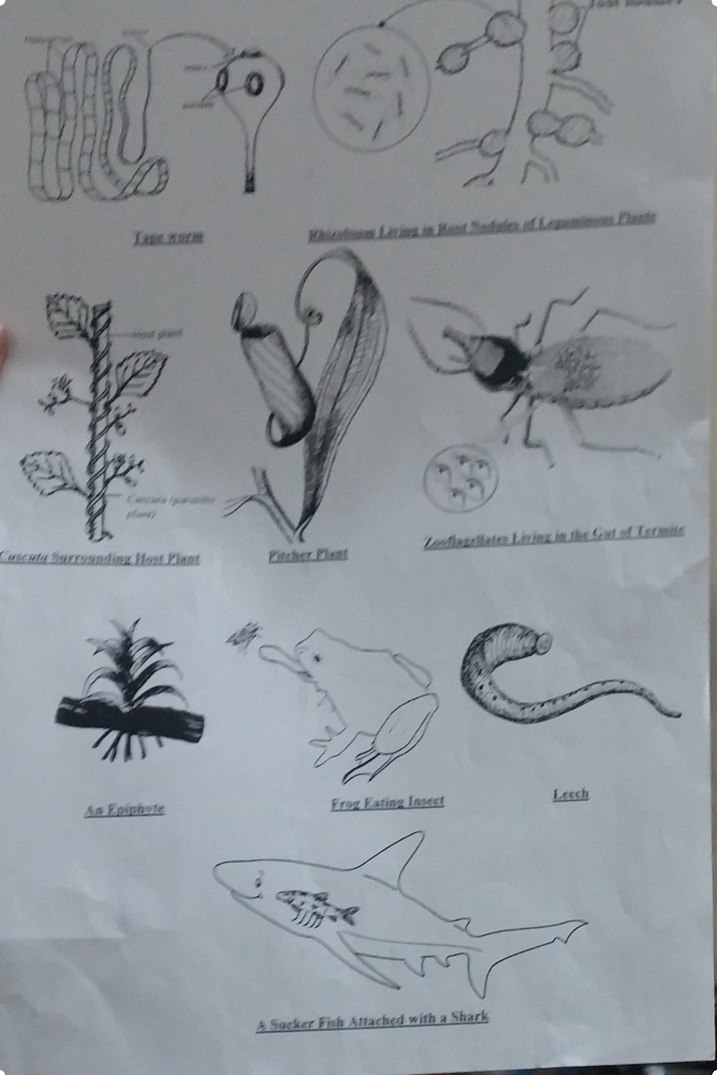} &
\includegraphics[width=0.20\textwidth]{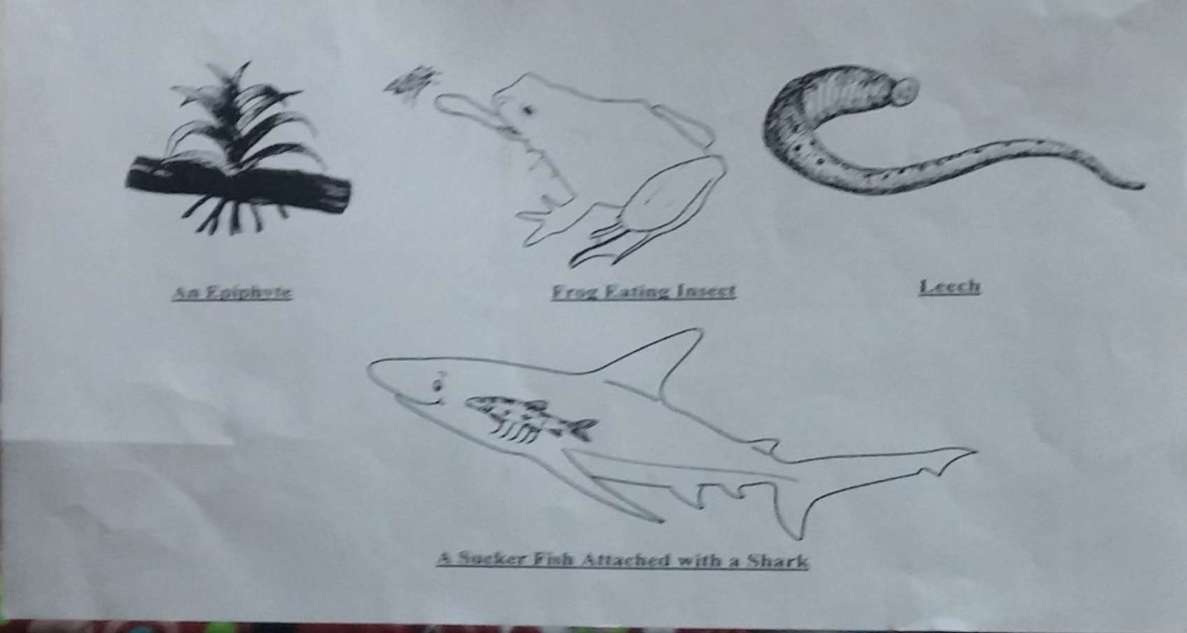} &
\includegraphics[width=0.20\textwidth]{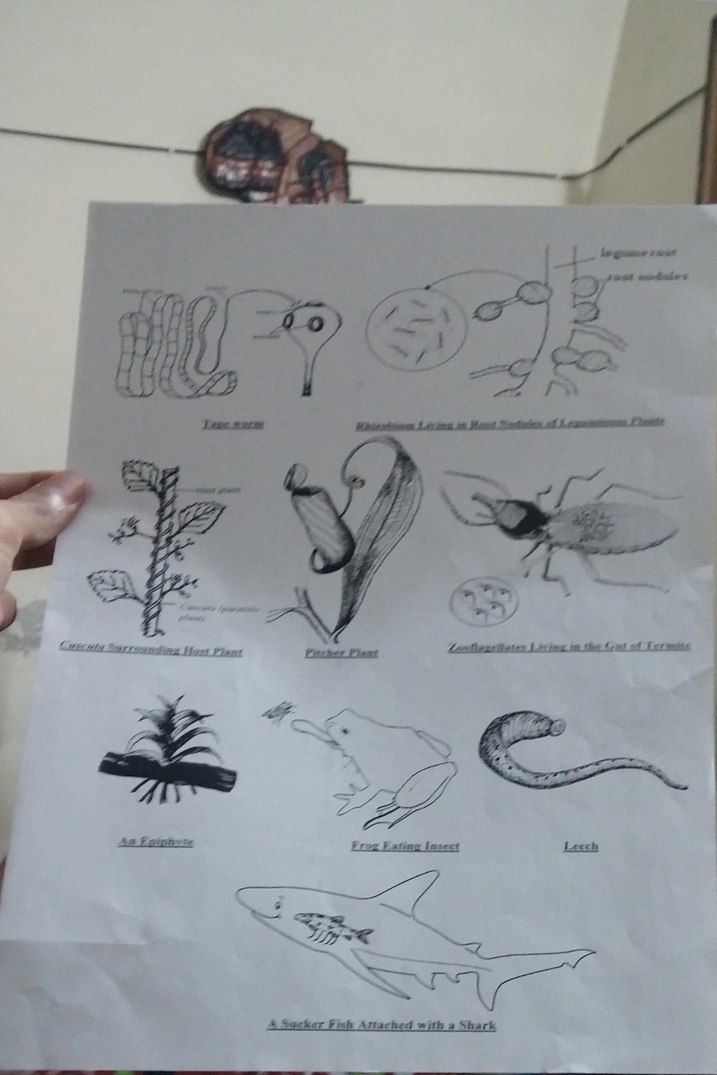} 
\end{tabular}
\caption{Visual comparison of results produced by our method and popular commercial softwares on test images. The proposed method gives better results over different range of photometric and geometric transformations and works independent of document underlying content. }
\label{figure:software_comp_2}
\end{center}
\end{figure*}
\subsection{Comparison with Commercial Software Applications}
\label{sec:software_apps}
Algorithms for rectifying perspectively distorted documents are also being used by many commercial software  applications for the purpose of optical character recognition and documents digitization. Here we compare our method with two popular commercial applications, \ie 
CamScanner\footnote{It \url{https://www.camscanner.com/} is one of the most famously used application  with around 50 million downloads, the most number of downloads for an android scanner application.} and ABBYY-Reader\footnote{\url{www.abbyy.com/en-apac/}}. 

We performed comparison with these commercial software at the three levels. At first level, we did the comparison via visual inspection of rectified images. CamScanner performed well for the cases where corners or edges of the documents were clearly visible and the documents could be distinguished from the background. In other cases where there were strong illumination artifacts, background clutter or the corners were not visible, the application failed to remove perspective distortion from the documents -- \cf \figrefs{software_comp}{software_comp_2} ($3_{rd}$ column) for more details. ABBYY-Reader gave good performance for the cases where documents edges were strong and documents could be easily differentiated from the background clutter. However, it failed in the cases where there were: (i) no textual line cues; (ii) strong illumination artifacts; or (iii) large scale geometric transformations in the captured documents. 

At second level, we first randomly sampled 400 images from our test set and passed them to CamScanner, ABBYY-Reader and our method for perspective correction. We then manually annotated true corner positions in these rectified documents and finally used these annotations to measure MDE \wrt ground truth. Compared to MDE of 21.5 and 20.6 pixels for ABBYY-Reader and CamScanner, our method achieves a MDE of 2.56 pixels on this dataset. 

At the third level, we compare the OCR performance of these methods on a pair of test datasets (named as Simple and Complex) generated from the high-resolution ground truth images of SmartDoc-QA dataset. The simple version of dataset was generated with least variations in photometric and geometric transformations, whereas the complex version includes the same level of variability (except motion blur) as included in our original synthetic dataset. \tabref{ocr-sdoc-syn} compare the performance of Tesseract OCR with different image rectification algorithms on these datasets. Our algorithm here once again consistently gives better performance than these competing methods. Although on the simple variant of dataset CamScanner is able to give comparable performance, however as the large variations are introduced both the CamScanner and ABBYY-Reader give much worse performance than our methods.

These results prove that our method is indeed a generic method and gives excellent results for document images captured under wide range of wild-settings.

\vspace*{-.75em}
\section{Conclusions}
\label{sec:conclusion}

In this paper, we have proposed a simple and efficient method to recover homography from the perspectively distorted document images. We have performed extensive experiments and shown that our proposed method gives excellent results in wide range of realistic image capturing settings. In comparison to earlier methods, our method works independent of documents contents and is fully automatic, as it does not require any manual input. 

Furthermore, for training deep networks, we have introduced a new synthetic dataset with warped camera captured documents that contains a large number of images compared to the present ones. Overall, following are the major findings of this study: (i) a rich dataset, even a synthetic one,  that records the true underlying real world distribution of problem plays a critical role in the overall performance of deep networks; (ii) in initial layers filters, large receptive fields are crucial for improved performance, also having large number of filters and convolutional layers are necessary to achieve state of the art performance for problem at hand; (iii) \loloss loss can be reliably used for regressing corner positions compared  to traditionally used \ltloss loss, however the overall difference in performance is not statistically significant. (iv) similar to \cite{detone2016deep}, we found that 4-points homography parameterization method works better than traditionally used $3\times3$ matrix representation and results in a stable loss function that gives state of the art performance. 

We are of the view that our method can become an integral part of complete document analysis pipeline. 

\clearpage
{\small
\bibliographystyle{ieee}
\bibliography{egbib}

\begin{thebibliography}{10}\itemsep=-1pt

\bibitem{abadi2016tensorflow}
M.~Abadi, A.~Agarwal, P.~Barham, E.~Brevdo, Z.~Chen, C.~Citro, G.~S. Corrado,
  A.~Davis, J.~Dean, M.~Devin, et~al.
\newblock Tensorflow: Large-scale machine learning on heterogeneous distributed
  systems.
\newblock {\em arXiv preprint arXiv:1603.04467}, 2016.

\bibitem{abdelmounaime2013new}
S.~Abdelmounaime and H.~Dong-Chen.
\newblock New brodatz-based image databases for grayscale color and multiband
  texture analysis.
\newblock {\em ISRN Machine Vision}, 2013, 2013.

\bibitem{baker2006parameterizing}
S.~Baker, A.~Datta, and T.~Kanade.
\newblock Parameterizing homographies.
\newblock {\em Robotics Institute, Carnegie Mellon University, Tech. Rep},
  2006.

\bibitem{bukhari2009dewarping}
S.~S. Bukhari, F.~Shafait, and T.~M. Breuel.
\newblock Dewarping of document images using coupled-snakes.
\newblock In {\em Proceedings of Third International Workshop on Camera-Based
  Document Analysis and Recognition, Barcelona, Spain}, pages 34--41. Citeseer,
  2009.

\bibitem{cimpoi14describing}
M.~Cimpoi, S.~Maji, I.~Kokkinos, S.~Mohamed, , and A.~Vedaldi.
\newblock Describing textures in the wild.
\newblock In {\em Proceedings of the {IEEE} Conf. on Computer Vision and
  Pattern Recognition ({CVPR})}, 2014.

\bibitem{deng2009imagenet}
J.~Deng, W.~Dong, R.~Socher, L.-J. Li, K.~Li, and L.~Fei-Fei.
\newblock Imagenet: A large-scale hierarchical image database.
\newblock In {\em Computer Vision and Pattern Recognition, 2009. CVPR 2009.
  IEEE Conference on}, pages 248--255. IEEE, 2009.

\bibitem{detone2016deep}
D.~DeTone, T.~Malisiewicz, and A.~Rabinovich.
\newblock Deep image homography estimation.
\newblock {\em arXiv preprint arXiv:1606.03798}, 2016.

\bibitem{dubrofsky2009homography}
E.~Dubrofsky.
\newblock {\em Homography estimation}.
\newblock PhD thesis, MS Thesis, UNIVERSITY OF BRITISH COLUMBIA (Vancouver),
  2009.

\bibitem{gupta2016synthetic}
A.~Gupta, A.~Vedaldi, and A.~Zisserman.
\newblock Synthetic data for text localisation in natural images.
\newblock {\em arXiv preprint arXiv:1604.06646}, 2016.

\bibitem{hartley2003multiple}
R.~Hartley and A.~Zisserman.
\newblock {\em Multiple view geometry in computer vision}.
\newblock Cambridge university press, 2003.

\bibitem{he2015delving}
K.~He, X.~Zhang, S.~Ren, and J.~Sun.
\newblock Delving deep into rectifiers: Surpassing human-level performance on
  imagenet classification.
\newblock In {\em Proceedings of the IEEE International Conference on Computer
  Vision}, pages 1026--1034, 2015.

\bibitem{hradivs2015convolutional}
M.~Hradi{\v{s}}, J.~Kotera, P.~Zemc{\'\i}k, and F.~{\v{S}}roubek.
\newblock Convolutional neural networks for direct text deblurring.
\newblock In {\em Proceedings of BMVC}, pages 2015--10, 2015.

\bibitem{jagannathanperspective}
L.~Jagannathan and C.~Jawahar.
\newblock Perspective correction methods for camera-based document analysis.

\bibitem{kingma2014adam}
D.~Kingma and J.~Ba.
\newblock Adam: A method for stochastic optimization.
\newblock {\em arXiv preprint arXiv:1412.6980}, 2014.

\bibitem{laina2016deeper}
I.~Laina, C.~Rupprecht, V.~Belagiannis, F.~Tombari, and N.~Navab.
\newblock Deeper depth prediction with fully convolutional residual networks.
\newblock {\em arXiv preprint arXiv:1606.00373}, 2016.

\bibitem{lampert2005oblivious}
C.~H. Lampert, T.~Braun, A.~Ulges, D.~Keysers, and T.~M. Breuel.
\newblock Oblivious document capture and real-time retrieval.

\bibitem{liang2008geometric}
J.~Liang, D.~DeMenthon, and D.~Doermann.
\newblock Geometric rectification of camera-captured document images.
\newblock {\em IEEE Transactions on Pattern Analysis and Machine Intelligence},
  30(4):591--605, 2008.

\bibitem{nayef2015smartdoc}
N.~Nayef, M.~M. Luqman, S.~Prum, S.~Eskenazi, J.~Chazalon, and J.-M. Ogier.
\newblock Smartdoc-qa: A dataset for quality assessment of smartphone captured
  document images-single and multiple distortions.
\newblock In {\em Document Analysis and Recognition (ICDAR), 2015 13th
  International Conference on}, pages 1231--1235. IEEE, 2015.

\bibitem{peng2015learning}
X.~Peng, B.~Sun, K.~Ali, and K.~Saenko.
\newblock Learning deep object detectors from 3d models.
\newblock In {\em Proceedings of the IEEE International Conference on Computer
  Vision}, pages 1278--1286, 2015.

\bibitem{quattoni2009recognizing}
A.~Quattoni and A.~Torralba.
\newblock Recognizing indoor scenes.
\newblock In {\em Computer Vision and Pattern Recognition, 2009. CVPR 2009.
  IEEE Conference on}, pages 413--420. IEEE, 2009.

\bibitem{yolo2015}
J.~Redmon, S.~Divvala, R.~Girshick, and A.~Farhadi.
\newblock You only look once: Unified, real-time object detection.
\newblock {\em arXiv preprint arXiv:1506.02640}, 2015.

\bibitem{shafait2007document}
F.~Shafait and T.~M. Breuel.
\newblock Document image dewarping contest.
\newblock In {\em 2nd Int. Workshop on Camera-Based Document Analysis and
  Recognition, Curitiba, Brazil}, pages 181--188, 2007.

\bibitem{simon2015correcting}
C.~Simon, I.~K. Park, et~al.
\newblock Correcting geometric and photometric distortion of document images on
  a smartphone.
\newblock {\em Journal of Electronic Imaging}, 24(1):013038--013038, 2015.

\bibitem{simonyan2014very}
K.~Simonyan and A.~Zisserman.
\newblock Very deep convolutional networks for large-scale image recognition.
\newblock {\em arXiv preprint arXiv:1409.1556}, 2014.

\bibitem{tieleman2012lecture}
T.~Tieleman and G.~Hinton.
\newblock Lecture 6.5-rmsprop: Divide the gradient by a running average of its
  recent magnitude.
\newblock {\em COURSERA: Neural Networks for Machine Learning}, 4(2), 2012.

\bibitem{ulges2004document}
A.~Ulges, C.~H. Lampert, and T.~Breuel.
\newblock Document capture using stereo vision.
\newblock In {\em Proceedings of the 2004 ACM symposium on Document
  engineering}, pages 198--200. ACM, 2004.

\bibitem{zhang2004restoration}
Z.~Zhang, C.~L. Tan, and L.~Fan.
\newblock Restoration of curved document images through 3d shape modeling.
\newblock In {\em Computer Vision and Pattern Recognition, 2004. CVPR 2004.
  Proceedings of the 2004 IEEE Computer Society Conference on}, volume~1, pages
  I--10. IEEE, 2004.

\bibitem{zwald2012berhu}
L.~Zwald and S.~Lambert-Lacroix.
\newblock The berhu penalty and the grouped effect.
\newblock {\em arXiv preprint arXiv:1207.6868}, 2012.

\end{thebibliography}
}

\end{document}